\RequirePackage{fix-cm}
\documentclass[smallextended]{svjour3}       
\smartqed  
\usepackage{graphicx}
\usepackage{listings}
\usepackage{comment}
\usepackage{graphicx}
\usepackage{epsfig}
\usepackage{array, colortbl}
\newcolumntype{P}[1]{>{\raggedright\arraybackslash}p{#1}}
\usepackage{listings}
\usepackage{epstopdf}
\usepackage{multirow}
\usepackage{rotating}
\usepackage{float}
\usepackage{balance}
\usepackage{scalefnt}
\usepackage[normalem]{ulem}
\usepackage[hang]{footmisc}
\usepackage{booktabs}
\usepackage{pbox}
\usepackage{ragged2e}
\usepackage{fancybox}
\usepackage{guehene}
\usepackage{chngpage} 
\usepackage{soul}
\usepackage[hyphens]{url}
\usepackage{hyperref}
\usepackage[T1]{fontenc} 
\usepackage{amsmath}
\usepackage[english]{babel}
\usepackage[english=american]{csquotes}
\usepackage[edges]{forest}
\usepackage{tikz}
\usepackage{ltablex}
\usepackage{xltabular}
\usetikzlibrary{trees}
\usepackage{algorithm}
\usepackage{algpseudocode}
\usepackage{multicol}

\usepackage[toc,page]{appendix}
\usepackage[authoryear,round]{natbib}

\usepackage{enumitem}

\usepackage[skins]{tcolorbox}
\tcbuselibrary{breakable}

\SetLabelAlign{center}{\hss#1\hss}
\tikzset{parent/.style={align=center,text width=2.5cm,rounded corners=2pt},
    child/.style={align=center,text width=2.5cm,rounded corners=2pt}
    }

\definecolor{dkgreen}{rgb}{0,0.6,0}
\definecolor{gray}{rgb}{0.5,0.5,0.5}

\makeatletter

\renewcommand\paragraph{\@startsection{paragraph}{4}{\z@}%
                    {-2.5ex\@plus -1ex \@minus -.2ex}
                                    {0.7ex \@plus .2ex}%
                                    {\normalfont\normalsize\bfseries}}
\makeatother

\newcommand*{\affmark}[1][*]{\textsuperscript{#1}}

\journalname{Automated Software Engineering}
\usepackage{caption}
\usepackage{subcaption}
\graphicspath{ {./images/} }
\begin{document}
\sloppy

\title{Data Cleaning and Machine Learning: A Systematic Literature Review  \thanks{This work is funded by the Fonds de Recherche du Quebec (FRQ), the Canadian Institute for Advanced Research (CIFAR), and the Natural Sciences and Engineering Research Council of Canada (NSERC).}
}


\author{Pierre-Olivier Côté\affmark[1] \and Amin Nikanjam\affmark[1] \and  Nafisa Ahmed\affmark[1] \and  Dmytro Humeniuk\affmark[1] \and  Foutse Khomh\affmark[1]}

\authorrunning{Côté et al.} 

\institute{\affmark[1] Polytechnique Montréal, Québec, Canada \\
            \email{\{pierre-olivier.cote, amin.nikanjam, nafisa.abdelmutalab-ali-ahmed@polymtl.ca,  dmytro.humeniuk@polymtl.ca, foutse.khomh\}@polymtl.ca}}

\date{Received: date / Accepted: date}

\maketitle

\begin{abstract}
\textbf{Context:} Machine Learning (ML) is integrated into a growing number of systems for various applications. Because the performance of an ML model is highly dependent on the quality of the data it has been trained on, there is a growing interest in approaches to detect and repair data errors (i.e., data cleaning). Researchers are also exploring how ML can be used for data cleaning; hence creating a dual relationship between ML and data cleaning. To the best of our knowledge, there is no study that comprehensively reviews this relationship.
\textbf{Objective:} This paper's objectives are twofold. First, it aims to summarize the latest approaches for data cleaning for ML and ML for data cleaning. Second, it provides future work recommendations. 
\textbf{Method:} We conduct a systematic literature review of the papers published between 2016 and 2022 inclusively. We identify different types of data cleaning activities with and for ML: feature cleaning, label cleaning, entity matching, outlier detection, imputation, and holistic data cleaning.
\textbf{Results:} We summarize the content of 101 papers covering various data cleaning activities and provide 24 future work recommendations. Our review highlights many promising data cleaning techniques that can be further extended.
\textbf{Conclusion:} We believe that our review of the literature will help the community develop better approaches to clean data. 


\end{abstract}

\section{Introduction}\label{sec:introduction}
Nowadays, Machine Learning (ML) integrates a growing number of industries, from transportation to healthcare and education. Recent applications of ML achieved performances similar to humans' performance on complex tasks, such as taking the bar exam \citep{GPT_bar_exam} or driving a car \citep{badue2021self, Gitnux_2023}. ML models are implemented as software components, and then integrated into other components in Machine Learning Software Systems (MLSSs). Such systems employ trained ML models to intelligently make decisions or generate output based on learned data-derived knowledge, and similar to any software systems, they need quality assurance \citep{gezici2022systematic}.

Behind much of the recent success of ML applications are large amounts of training data and powerful computing infrastructure \citep{roh2019survey, whang2021data}. While a major part of the research in ML is spent on developing better modeling techniques \citep{Ng_2021}, data preparation often is the most arduous and time-consuming task for practitioners. Indeed, researchers have reported that data scientists sometimes spend over 80\% of their time preparing data \citep{whang2021data, neutatz2021cleaning, deng2017data, agrawal2019cloudy}. As a consequence, there is an emerging trend, referred to as Data-Centric AI (DCAI) to steer research to focus on improving datasets instead of models for ML problems. In the last few years, various initiatives aimed at stimulating involvement in the DCAI trend have been proposed. For example, in August 2021 the first DCAI competition \citep{DCAI} was organized. During this competition, participants were challenged to improve a model's performance by only modifying the dataset it was trained on. Following this competition, a group of researchers released the DataPerf benchmark \citep{mazumder2022dataperf}, a suite of data-centric benchmarks for different data tasks such as data cleaning and data debugging. 

Amongst the different steps forming data preparation is a central task that consists of detecting and removing data errors, namely data cleaning \citep{zha2023data}. Previous studies have often overlooked it assuming that the datasets used in their experiments are devoid of errors. However, as reported by a previous study \citep{https://doi.org/10.48550/arxiv.1911.00068}, large public datasets considered for a long time to be error-free such as ImageNet \citep{russakovsky2015imagenet} contain erroneous labels. As a consequence, there is an increasing interest in developing data-cleaning approaches to improve machine learning performance. Data cleaning is a problem tackled by the database community for a long time \citep{rahm2000data, fox1999maintaining, mayfield2010eracer} and, recently, researchers started studying how ML can be leveraged to clean data.

In this study, we summarize the latest approaches in Data Cleaning for ML (DC4ML) and ML for Data Cleaning (ML4DC). For simplicity, we refer to the union of DC4ML and ML4DC as data cleaning and ML (DC\&ML). To reach that goal, we adopt the Systematic Literature Review (SLR) approach, which differs from traditional literature reviews by its strict and rigorous methodology. The SLR methodology ensures that the literature review is exhaustive and devoid of biases when collecting and summarizing information \citep{kitchenham2004procedures}. With the increasing amount of work published on DC\&ML (see Section \ref{sec:results:stats}), we believe that there is a need to summarize the latest approaches in the field. A replication package of our study is available on our public GitHub repository \citep{replicationpackage}. In total, we summarize the content of 101 papers and provide 24 future work directions. Our review can serve as a basis for researchers and practitioners to understand state-of-the-art DC\&ML approaches and to contribute to the field.

\textbf{The rest of the paper is organized as follows}. In Section \ref{sec:methodology}, the methodology followed is described. Then, in Section \ref{sec:results:tax}, the taxonomy selected to structure our review of the literature is presented. Next, statistics about the papers included in our study are detailed in Section \ref{sec:results:stats}. The review of the literature is presented in Section \ref{sec:review} and future research directions are proposed in Section \ref{sec:future-works}. Finally, we reflect on our work and discuss threats to validity in Section \ref{sec:threats}. The conclusion of the paper is presented in Section \ref{sec:conclusion}.

\section{Background}

Recently, Machine Learning Software Systems (MLSSs) have seamlessly integrated into our everyday lives, manifesting in various forms such as recommendation systems, speech recognition, and face detection. Typically, an MLSS ingests data and utilizes ML models to autonomously make intelligent decisions based on learned patterns, associations, and data-derived knowledge \citep{10.1145/3487043}. These ML models function as software components, interacting with other software components within MLSSs. Similar to conventional software systems, MLSSs necessitate quality assurance, particularly given their increasing significance in today's society. The ramifications of erroneous or subpar decisions by such systems can range from system malfunctions to substantial financial losses or even endangerment to human life \citep{foidl2019risk}. Assessing the quality of MLSSs presents a formidable challenge and is currently a focal point of research \citep{gezici2022systematic,10.1145/3487043}. Recent research on MLSS quality underscores the importance of encompassing various dimensions of quality \citep{10.1145/3487043,studer2021towards}, not solely focusing on prediction accuracy.

The quality of an MLSS is contingent upon numerous factors. In MLSSs, both the code (representing the implemented model) and the training data play pivotal roles in shaping system behavior \citep{felderer2019testing,whang2023data}. Compared to traditional Software Engineering (SE), the logic employed by ML models is learned from the data instead of being coded, so data quality is key to achieving good model quality. Consequently, ensuring high data quality and validating the integrity of data becomes imperative requirements in such data-dependent systems as the quality of the trained model impacts the overall quality of the system \citep{BRAIEK2020110542}. Data cleaning is the task of detecting or repairing corrupted, duplicate, incomplete, inaccurate, or noisy records from the dataset, to improve the quality of the dataset. This improves the quality of any potential MLSS that is trained on/uses the cleaned dataset.

Because of the importance of data quality for MLSS, there is an increasing interest in developing data-cleaning approaches to improve ML performance. Tools to address label errors in datasets, such as the platform proposed by snorkel.ai \citep{Team_2024}, are gaining traction in the industry. Large companies are now proposing their own tools to improve dataset quality, such as IBM’s tool “Data Quality for AI” \citep{10.1145/3493700.3493774}. In parallel, in the research community, there is a growing number of studies trying to summarize the latest advancement in this fast-moving field \citep{neutatz2021cleaning, 10.1145/3506712, thirumuruganathan2020data, roh2019survey, chai2022data}.

While data cleaning plays an increasingly important role in ML, the latter can also be used to enhance data cleaning processes. Data cleaning is a problem initially tackled by the database community for a long time \citep{rahm2000data, fox1999maintaining, mayfield2010eracer} and, recently, researchers started studying how ML can be leveraged to clean data. Integrating ML into data cleaning systems presents an attractive opportunity for researchers and practitioners in the field since it has the capacity to simplify current cleaning processes. Instead of manually devising a set of cleaning rules and being burdened with its maintenance, ML offers the possibility of automatically learning these rules from a dataset. Additionally, previous data cleaning efforts can be easily integrated into any ML model as model features, as mentioned in \citet{10.1145/3506712}. Finally, the use of ML for data cleaning enables data cleaning systems to benefit from the advancements in the field of ML and opens up many new research opportunities. Thus, as much as ML can benefit from data cleaning, data cleaning may also profit from ML.

\section{Methodology}\label{sec:methodology}

\begin{figure}[t]
 \centering
\includegraphics[width=0.85\textwidth]{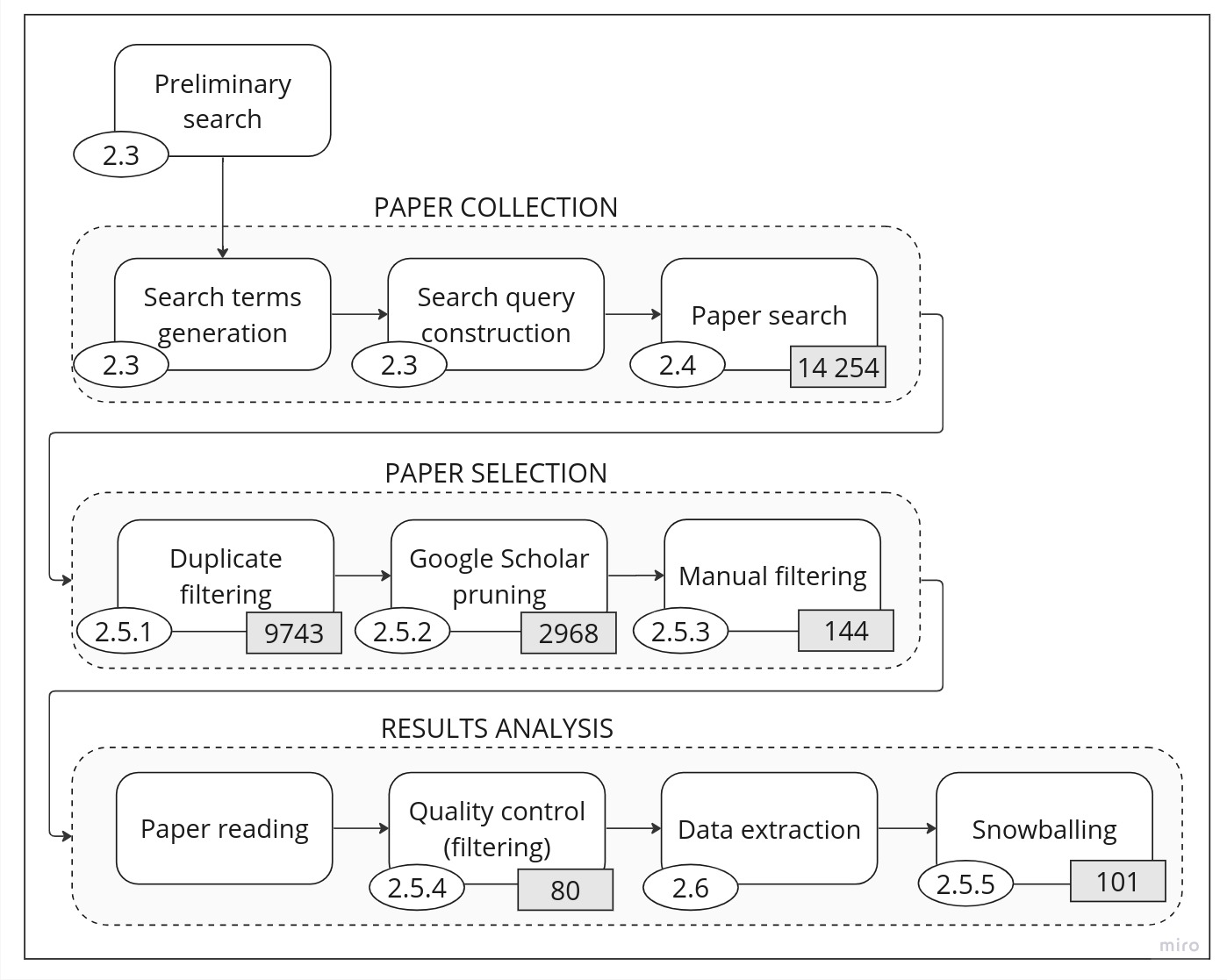}
 \caption{Outline of our methodology}
 \label{fig:overview}
\end{figure}

In this section, we describe the methodology we followed to conduct our study. We provide an outline of it in Figure \ref{fig:overview}. Our methodology is divided into three phases: (1) paper collection, (2) paper selection, and (3) results analysis. We collected papers on 17 October 2022. For each step in the figure, we indicate (1) the section where this step is explained (in the lower left corner of the box) and (2) the number of papers left after the step (in the lower right corner of the box). To conduct our study, we followed the guidelines defined by \citet{kitchenham2004procedures} for systematic literature reviews. The author described best practices for conducting systematic reviews in SE and is based on existing guidelines from medical research, where empirical research methods are mature and rigorous. As recommended by the authors \citet{kitchenham2004procedures}, we started by defining our research question (Section 3.1). We then generated our search query (Section 3.3) and collected papers from academic databases (Section 3.4). We filtered out papers that were irrelevant to our study based on selected criteria and quality control questions (Section 3.5). Additional relevant papers were found using snowballing (Section 3.6). To ensure an accurate extraction of data (Section 3.7), we designed and used data extraction forms, following \citet{kitchenham2004procedures} recommendations.

\subsection{Research Questions}\label{sec:RQs}
As we mentioned earlier, the goal of this study is to conduct a systematic review of DC\&ML. Thus, we define the following research questions.

\begin{description}
    \item[RQ1:] \textbf{What are the latest data cleaning techniques in DC\&ML?} As mentioned before, a significant portion of practitioners' time when building machine learning systems is spent on data. In parallel, ML has made significant progress in the last ten years and is applied to a growing number of applications. Applying ML to data cleaning has the potential of reducing the amount of time spent on data by practitioners, as well as improving the accuracy of data cleaning processes. Conversely, data cleaning can play an important role in improving a model's performance. By answering this RQ, we provide researchers with an understanding of the current data cleaning approaches in DC\&ML. 
    
    \item[RQ2:] \textbf{What are future research opportunities in data cleaning and ML?} In this RQ, our objective is to identify and highlight promising research directions in the field of data cleaning with machine learning.
\end{description}

\subsection{Scope} \label{Methodology:scope}
For a long time, data cleaning was mainly applied to tabular data (i.e., data stored in relational databases), and surveys on data cleaning focused primarily on this type of data \citep{10.1145/2882903.2912574}. Nowadays, data cleaning techniques for various data types such as images \citep{ponzio2021w2wnet}, text \citep{klie2022annotation}, graph data \citep{https://doi.org/10.48550/arxiv.2202.09747}, and sound \citep{9721724} are developed. The strategy used to clean data in these techniques may significantly vary from one data type to another because of the particularities of each data type. In our survey, we focused on the data types that are the most commonly processed by ML applications in practice. We used comprehensive surveys and books from the ML community to identify them \citep{sarker2021deep, pouyanfar2018survey, Goodfellow-et-al-2016, cote2023quality}. Hence, we limited ourselves to tabular, image, and text data.

\subsection{Search Terms Selection} \label{subsubsection:search_terms}
\lstset{
    language=Pascal,
    showstringspaces=false,
    columns=flexible,
    basicstyle={\scriptsize\ttfamily},
    keywordstyle=\color{blue},
    commentstyle=\color{dkgreen},
    breaklines=true,
    breakatwhitespace=true,
    tabsize=4,
}
\begin{figure*}[t]
\centering
        \begin{lstlisting}

("machine learning" OR "deep learning" OR "neural network" OR "neural networks" OR "reinforcement learning" OR supervised OR unsupervised)   
AND 
(
  ( "data cleaning" OR "data cleansing" OR "data scrubbing") 
  OR ("data repairing" OR "data repair" OR "error repair" OR "error repairing") 
  OR ("confident learning" OR "label cleaning")
  OR ("error detection" AND (tab* OR cell* OR row* OR image* OR text*))
)
        \end{lstlisting}
    \caption{The search query used to fetch papers from all academic databases used in our study except Google Scholar.}
 \label{fig:search_pattern}
\end{figure*}

As one of the first steps of the SLR, we conducted a preliminary search to help us identify the most relevant keywords to our study, as recommended by \citet{kitchenham2004procedures}. Hence, we read existing survey of the literature \citep{ilyas2019data, neutatz2021cleaning, 10.1145/3506712, thirumuruganathan2020data, roh2019survey, whang2023data, chai2022data} and highly cited works (e.g., \citet{rekatsinas2017holoclean, krishnan2017boostclean}). Then, we defined a comprehensive search query to gather the relevant literature from academic databases. Our final search query is described in Figure \ref{fig:search_pattern}. In the following, we describe how we selected keywords based on the main topics of our review: `ML' and `data cleaning'. Then, we explain how we adapted our query for one of the academic databases we used, Google Scholar (GS). Contrary to other databases, GS had peculiarities that forced us to modify our search query.

\begin{itemize}
    \item \textit{Machine learning}: ML is a field devoted to enabling machines to learn from data \citep{Wikipedia_2023_ML}. We leveraged previous SLRs \citep{tambon2022certify, KANG2020106773, AZEEM2019115} on different topics in ML to build an initial list of keywords for ML. Our final list of terms is: \textit{machine learning}, \textit{deep learning}, \textit{neural network}, \textit{reinforcement learning}, \textit{supervised} and \textit{unsupervised}. 
    
    \item \textit{Data cleaning}: Data cleaning refers to the activity of detecting and repairing errors in data \citep{ilyas2019data}. We used the term \textit{data cleaning} and its synonyms \textit{data cleansing, data scrubbing} to conduct our searches. Additionally, we added keywords for both processes of data cleaning: error detection and error repair. For the former one, we used the keywords: \textit{data repairing, data repair, error repairing, error repair}. For the latter one, we only used one keyword: \textit{error detection}. After running the search query with the aforementioned keywords, we initially obtained a lot of irrelevant papers in the results. The reason is that error detection is a very broad topic that is not only limited to the detection of errors in data\footnote{For example, "A modular edge-/cloud-solution for automated error detection of industrial hairpin weldings using convolutional neural networks" is a paper that was included in the results but not relevant to our study since it is not a data cleaning approach.} Thus, in an attempt to filter out studies that do not propose an approach to detect data errors, we limited the results to papers that are also mentioning a data type. As mentioned in Section \ref{Methodology:scope}, we are focusing on three common data types: tabular, text, and image data. Hence, we used the following terms to limit the papers on error detection: \textit{tab*}, \textit{cell*}, \textit{row*}, \textit{image*} and \textit{text*}. Finally, because we are interested in data cleaning for ML (and ML datasets have labels), we added the following keywords: \textit{label cleaning} and \textit{confident learning}. 
\end{itemize}

\subsubsection{Adaptations for Google Scholar} \label{sec:search_terms:gs}
As opposed to other academic databases, GS has many peculiarities that require us to adapt our search pattern. Our final search pattern can be visualized in Figure \ref{fig:search_pattern:GS}. Below, we describe GS's peculiarities and explain how we adapted the query.
\begin{enumerate}
    \item All queries are truncated to 256 characters. 
    \item Parentheses are not a priority operator.
    \item \textit{OR} operators have a higher precedence than \textit{AND}.
    \item \textit{AND} operators are implicitly added between terms that are not operators (\textit{e.g.,} terms that are not \textit{OR} or \textit{AND}). Hence, the search query \textit{data cleaning} is equivalent to \textit{data AND cleaning}.
\end{enumerate}
In order to address the first constraint, we split the query in two and ran each part separately. For the same reason, we also removed any \textit{AND} operators, since they are redundant and added automatically (constraint 4). As parenthesis do not have their usual effect (constraint 2), we also removed them from our query. This prevents any nested Boolean operator. Hence, we removed the conjunction we described in Section \ref{subsubsection:search_terms} to limit the number of results on error detection.  

\subsection{Paper Collection}\label{Methodology.paper_search}
In the following, we describe how we collected papers from academic databases using the query described in Subsection \ref{subsubsection:search_terms}. Then, we describe the separate paper collection process we followed for GS, because of the limitations of that database. Similarly to the other works \citep{tambon2022certify, KANG2020106773, ALSOLAI2020106214}, we conducted the search in the following academic databases:
\begin{itemize}
    \item Google Scholar\footnote{https://scholar.google.com/}
    \item Engineering Village\footnote{https://www.engineeringvillage.com}
    \item Web of Science\footnote{https://webofknowledge.com}
    \item Science Direct\footnote{https://www.sciencedirect.com}
    \item ACM Digital Library\footnote{https://dl.acm.org}
    \item IEEE Xplore\footnote{https://ieeexplore.ieee.org}
\end{itemize}
As shown by the recent publications  
\citep{neutatz2021cleaning, 10.1145/3506712}, DC\&ML is a relatively new topic. Hence, when running our search query, we limited the results to papers published between 1 January 2016 and 17 October 2022 (the later date being when we ran the query). On Engineering Village, we used the duplicate removal feature, by keeping the results from Compendex over Inspec.

\lstset{
    language=Pascal,
    showstringspaces=false,
    columns=flexible,
    basicstyle={\scriptsize\ttfamily},
    keywordstyle=\color{blue},
    commentstyle=\color{dkgreen},
    breaklines=true,
    breakatwhitespace=true,
    tabsize=4,
}
\begin{figure*}[t]
\centering
        \begin{lstlisting}

"machine learning" OR "deep learning" OR "neural network" OR "reinforcement learning" OR "supervised" OR "unsupervised" 
"confident learning" OR "label cleaning" OR "error detection"

"machine learning" OR "deep learning" OR "neural network" OR "reinforcement learning" OR "supervised" OR "unsupervised" 
"data cleaning" OR "data cleansing" OR "data scrubbing" OR "data repairing" OR "data repair" OR "error repair" 
        \end{lstlisting}
    \caption{The search query used to fetch papers from Google Scholar.}
 \label{fig:search_pattern:GS}
\end{figure*}

\subsubsection{Adaptations for Google Scholar}
Unlike other databases, GS does not provide a way to export records in bulk, which discourages mining techniques. Thus, to alleviate the task of collecting paper, we used the tool ``Publish or Perish''\footnote{Harzing, A.W. (2007) Publish or Perish, available from https://harzing.com/resources/publish-or-perish}. 
Similarly to \citet{tambon2022certify}, we fetched, for each year, the 1000 most relevant records to our search query according to GS.
As mentioned in Section \ref{sec:search_terms:gs}, our initial query had to be divided into two parts because of GS's limitations. Thus, we separately ran both queries, for every year for a period of six years, and collected 1000 papers per year, for a total of 12 000 papers (2 queries $\times$ 6 years $\times$ 1000 papers = 12 000 papers).

\subsection{Paper Selection} \label{sec:paper_selection}
In this section, we explain the steps we followed to filter out irrelevant results. Table \ref{tab:raw_papers} shows the number of papers obtained from each academic database prior to any kind of filtering. We provide the list of papers retained after each step of the paper selection process in our replication package \citep{replicationpackage}.

\begin{table}[t]
	\centering
	\small
	\caption{Number of papers retrieved from the academic databases.}
	\begin{tabular}{ | l | r | }
		\hline
		\textbf{Database} & \textbf{Number of papers} \\ \hline	
 		Google Scholar & 12,000 \\ \hline
		Engineering village & 1224 \\ \hline
		Web of Science & 511 \\ \hline
		Science Direct & 135 \\ \hline
		ACM Digital Library & 89 \\ \hline
		IEEE Xplore & 295 \\ \hline
		Total & 14,254 \\ \hline
	\end{tabular}
	\label{tab:raw_papers}
\end{table}

\subsubsection{Duplicate Filtering}
We uploaded all the titles of papers obtained from the various databases in Zotero\footnote{https://www.zotero.org/}. We leveraged the duplicate removal functionality of the tool and found 3,742 duplicates in GS' papers and 769 in the results of the other databases. Thus, after this step, 8,258 papers from GS were retained, and 1,485 for the other databases.

\subsubsection{Google Scholar Pruning}
While GS returned a lot of results, a considerable amount of them are not relevant to our study, for two reasons. First, a lot of irrelevant papers are included because of the search engine's limitation which we describe in Section \ref{subsubsection:search_terms}. Second, GS only searches keywords in the full text of the paper or in the title. Thus, papers irrelevant to our study that mentioned data cleaning and ML in the papers' body (e.g., as key/index works) matched our search query. While we could not address the first issue in any other way than manually inspecting and then filtering papers, we devised an approach for the second issue. We managed to only keep papers whose title, abstract, and author keywords matched the search query, similar to what was done by a similar study \citep{tambon2022certify}. We did so using a Python script we implemented which is available in our replication package \citep{replicationpackage}. After this step, 1,862 results from GS remained. We merged this set of papers together with the 1,485 papers coming from the other databases. We removed duplicated papers from the merged set, resulting in a total of \textbf{2,968 papers}.

\subsubsection{Inclusion and Exclusion Criteria}\label{inclusion-exclusion-criteria}
From this step onward, papers were filtered manually. One author went through the whole set of papers and filtered out the ones irrelevant to our study. A second author verified the resulting set of papers. Any leftover duplicate was removed. For this step, the following inclusion and exclusion criteria were considered:

\bigskip
\underline{Inclusion criteria}:
\begin{itemize}
    \item Papers proposing an approach to detect or fix errors in an ML dataset.
    \item Papers presenting an approach to detect or fix errors in data while leveraging ML techniques.
\end{itemize}

\medskip
\underline{Exclusion criteria}:
\begin{itemize}
    \item Papers whose data cleaning approach is specific to their problem (i.e., the technique does not generalize).
    \item Papers on data cleaning for other data types than tabular, image, or text data, such as sound data.
    \item Papers on detecting or fixing any type of attacks on an ML model through data (i.e., data poisoning and adversarial examples).
    \item Papers that can not be accessed free of charge through our institution's subscriptions.
    \item Secondary works (e.g., surveys and review papers), research proposal, workshop, letters, undergrad theses, or position papers.
    \item Papers that are not written in English.
    \item Papers on text data cleaning for other languages than English.
\end{itemize}

After this step, \textbf{124 papers remained}.

\subsubsection{Quality Control Assessment} \label{quality-control}
While most papers are relevant to our study, not all of them are able to answer our RQs. Thus, based on a previous study \citep{tambon2022certify}, we defined a set of questions to filter out papers with low quality. We also defined a second set of questions to filter out any remaining paper that is irrelevant to our study. \\

\bigskip
Quality control questions:
\begin{enumerate}[leftmargin=4em]
    \item[Q1.1] Is the objective of the research clearly defined?
    \item[Q1.2] Is the context of research clearly defined?
    \item[Q1.3] Does the study bring value to academia or industry?
    \item[Q1.4] Are the findings clearly stated and supported by results?
    \item[Q1.5] Are limitations explicitly mentioned and analyzed?
    \item[Q1.6] Is the methodology clearly defined and justified?
    \item[Q1.7] Is the experiment clearly defined and justified?
\end{enumerate}

\medskip
Relevance control questions:
\begin{enumerate}[leftmargin=4em]
    \item[Q2.1] Does the paper present an approach to detect or fix errors in an ML dataset?
    \item[Q2.2] Does the paper present an approach to detect or fix errors in data using ML techniques?
\end{enumerate} 

Each paper is considered by a reviewer who assigns it a score based on how well it complies with the questions. Papers under a certain threshold are considered by a second reviewer who repeats the process followed by the first reviewer. If the second reviewer's score is also under the threshold, the paper under consideration is excluded from the study. However, if the second reviewer assigns a score above the threshold, the reviewers have to discuss to decide whether the paper should be included or not in our study. If they are unable to reach a consensus, a third reviewer makes the final decision.

In order to evaluate a paper, reviewers rated on a scale from 0 to 2 how much they agreed with the quality and relevance control questions. For the quality control questions (i.e., Q1.X), a paper is excluded by a reviewer if the sum of the 7 questions is lower than 7. For the relevance control question, a paper is excluded if it has a score of zero for both questions (i.e., Q2.1 and Q2.2).

As a result, we excluded 44 papers. Our ratings are available in our replication package \citep{replicationpackage}. After this step, \textbf{80 papers remained}.

\subsection{Snowballing} \label{met:snowballing}
Snowballing refers to the act of discovering new papers through paper references \citep{wohlin2014guidelines}. In this study, to read relevant studies we miss, we include any paper cited by at least two papers from the filtered list of papers from Section \ref{quality-control}. We partially automated this process with a script available in our replication package \citep{replicationpackage}. In this script, we used the Python library Scholarly \citep{cholewiak2021scholarly} to fetch paper citations from GS. We evaluate the inclusion and exclusion criteria from Section \ref{inclusion-exclusion-criteria} along with the control assessment questions from Section \ref{quality-control} before including a snowballed paper in our study. After this step, we found 21 new papers.  Hence, in total, we collected \textbf{101 papers for examination} in our study.

\subsection{Data extraction}\label{sec:data_extraction}
In order to reason efficiently over the large amount of information contained in the set of papers we read, we extracted the following information from each paper:
\bigskip

General information
\begin{itemize}
    \item Title
    \item URL to the paper
    \item Authors
    \item Year
    \item Publication venue
\end{itemize}

\medskip
Questions
\begin{itemize}
    \item What does the data cleaning approach try to fix (e.g. missing values, dirty data, inconsistent data, etc.)?
    \item What are the datasets used in the experiment?
    \item What are the ML models used in the experiment?
    \item What are the quality measurements (i.e., how is the data cleaning approach evaluated)?
    \item Against what is the approach compared (e.g., against baselines, other approaches, etc.)?
    \item What is the performance of the approach, in terms of ML performance (e.g., accuracy, F1-score, etc.)?
    \item What is the performance of the approach, in terms of engineering aspects (e.g., resource consumption, time to completion, etc.)?
    \item What are the limitations of the approach?
\end{itemize}

Additionally, reviewers are asked to summarize the papers they have reviewed, in order to give a general picture of the study and to highlight any other interesting information. All extracted data is available in our replication package \citep{replicationpackage}.

\section{Selected Taxonomy}\label{sec:results:tax}

\begin{figure}
      \centering
      \scalebox{.75}{
      \begin{forest}
                for tree={
                    grow'=east,
                    forked edges,
                    draw,
                    rounded corners,
                    node options={
                        align=center },
                    text width=3cm,
                    anchor=west,
                    l sep=1.5cm,
                }
                [DC\&ML \textbf{(101)}, fill=white!25, parent,
                [Feature Cleaning \textbf{(36)}, for tree={fill=red!25, child},
                [Model-based \textbf{(19)}, for tree={fill=red!25, child},
                    [ Ensemble-based \textbf{(7)}]
                    [ Transformer-based \textbf{(4)} ]
                    [ Autoencoder-based \textbf{(3)} ]
                ]
                [ Error prioritization \textbf{(3)} ]
                [ Data cleaning rule generation \textbf{(2)} ]
                ]
                [Label Cleaning \textbf{(32)},  for tree={fill=blue!25,child}
                [Uncertainty-based \textbf{(10)}]
                [Loss-based \textbf{(5)}]
                [Counterfactual \textbf{(9)}]
                [Outlier-based \textbf{(3)}]
                ]
                [Entity Matching \textbf{(20)},  for tree={fill=green!25, child}
                [Token comparison \textbf{(2)} ]
                [Latent space comparison \textbf{(11)},  for tree={fill=green!25, child}
                    [Token-level comparison \textbf{(4)}]
                    [Attribute-level comparison \textbf{(6)}]
                    [Record-level comparison \textbf{(1)}]
                ]
                [Learned comparison \textbf{(4)}]
                ]
                [Outlier detection \textbf{(8)},  for tree={fill=yellow!35, child}
                [Statistic-based \textbf{(1)}]
                [Distance-based \textbf{(2)}]
                [Model-based \textbf{(2)}]
                ]
                [Imputation \textbf{(3)},  for tree={fill=brown!40, child}
                ]
                [ Holistic Data Cleaning \textbf{(4)}, for tree={fill=purple!35, child},
                [,phantom]
                ]
                ]     
                ]
            \end{forest}} 
\caption{Overview of the approaches discussed in this paper}
\label{fig:taxonomy}
\end{figure}
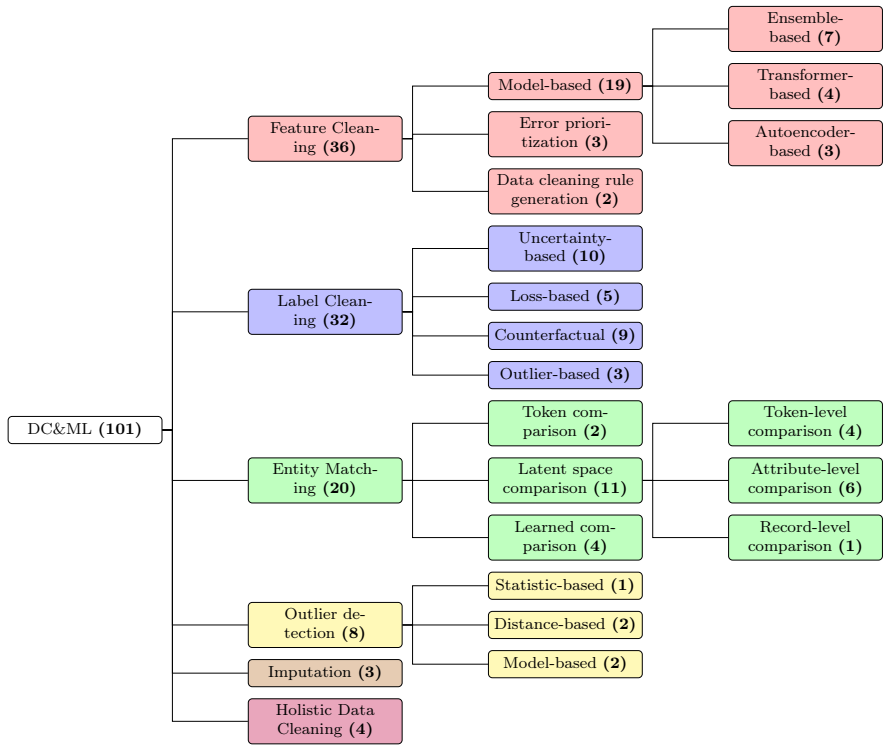

As mentioned in Section 1, data cleaning consists of detecting and removing data errors. To categorize the collected papers, similar to existing studies \citep{ilyas2019data, 10.1145/3506712}, we first grouped data cleaning activities described in the papers based on the type of error they try to address (e.g., incorrect feature value, duplicates, etc.). The data error addressed for each data cleaning activity is highlighted in Table \ref{data_error_type}. Then, we grouped papers in each main category based on the strategy employed to clean data. For example, token-comparison approaches (Section \ref{em:token_comp}) detect duplicate records by comparing their values, while latent space comparison approaches (Section \ref{em:latent}) compare the embeddings that represent the records. Figure \ref{fig:taxonomy} illustrates the identified taxonomy. Bold font is used in the figure to indicate the number of papers included in a category. The number of papers in a category may not be equal to the sum of the number of papers in its subcategories because a paper may be classified to zero or many subcategories of a category it belongs to. 

\begin{table}[h]
\centering
\begin{tabular}{|c|c|} 
 \hline
 \textbf{Data cleaning activity} & \textbf{Data error type} \\ 
 \hline
    Feature cleaning & Incorrect feature values \\
    \hline
    Label cleaning &Incorrect label values \\
    \hline
    Entity matching & Duplicate records \\
    \hline
    Outlier detection & Out-of-distribution records \\
    \hline
    Imputation & Missing values \\
    \hline
    Holistic data cleaning  & More than one error type at the same time \\ 
 \hline
\end{tabular}
\caption{Data error type addressed per data cleaning activity}
\label{data_error_type}
\end{table}

In the following, we introduce the different types of data cleaning activities, which correspond to the categories on the first level of the taxonomy in Figure \ref{fig:taxonomy}. The other categories will be introduced later in their respective subsections during the review (in Section \ref{sec:review}).

\begin{itemize}
    \item \textbf{Feature cleaning}: Feature cleaning approaches try to detect or repair errors in the features of a record. It is a problem that has been addressed by the database community for a long time \citep{rahm2000data, fox1999maintaining, mayfield2010eracer} and is generally loosely referred to as data cleaning. It should be noted that data cleaning is not limited to feature cleaning and that other activities are also considered to be data cleaning activities (e.g., entity matching or outlier detection \citep{ilyas2019data}). Thus, we use the term ``feature cleaning'' instead of ``data cleaning'' to refer to that data cleaning activity. Feature errors may happen for a large variety of reasons. For example, a noun may be misspelled by the person collecting data \citep{cote2023quality}.
    \item \textbf{Label cleaning}: Similar to feature cleaning, label cleaning approaches try to detect or repair label errors. Label errors may happen for a variety of reasons. For example, a record might be mislabeled because of a lack of domain knowledge \citep{sambasivan2021everyone}. It should be noted that label cleaning is different from label-noise robust learning \citep{9729424}. While both approaches’ main objective is to improve the performance of ML models, the strategy followed to reach that goal differs.  Instead of fixing the data errors, label-noise robust learning approaches modify the learning algorithms in order to maintain the performance of ML models in the presence of label noise \citep{https://doi.org/10.48550/arxiv.1911.00068}. Thus, label-noise robust learning is not a form of data cleaning nor is it label cleaning. We expand on label-noise robust learning in Section \ref{appendix:robust-learning} and compare it with label cleaning.
   \item \textbf{Entity matching}: Entity matching refers to the process of finding tuples in a database that refer to the same real-world entity \citep{ilyas2019data}. Duplicates can be introduced in a dataset for many reasons. For example, a customer could be recorded multiple times in a database because the customer used different names at checkout \citep{ilyas2019data}.
    \item \textbf{Outlier detection}: An outlier can be defined as an observation that deviates so much from the other records in a dataset that it arouses suspicions that it has been generated by a different process \citep{hawkins1980identification}. Outliers may hinder an ML model's performance if they have been generated by a different process. Hence, outliers may be purged from a dataset during data cleaning. 
    \item \textbf{Imputation}: Imputation refers to assigning a value to a record's feature whose value is missing \citep{Wikipedia_2023_imputation}. Having missing values is a frequent problem \citep{alimohammadi2022performance} and is often experienced when data collection is manual \citep{lakshminarayan1999imputation}. An ML model can only ingest complete records; thus records with missing values have to be imputed otherwise they are removed from datasets.
    \item \textbf{Holistic data cleaning}: We use the term ``holistic data cleaning'' to refer to the approaches that try to clean more than one type of error at a time. For example, \citet{berti2019learn2clean} designed an RL agent that learns to clean a dataset using a set of data cleaning tools that can clean different types of errors.
\end{itemize}

In light of the identified taxonomy, we devise sub-RQs as follows to answer our RQ1 on recent data cleaning techniques in DC\&ML:
\begin{itemize}[leftmargin=4.5em]
    \item[RQ1.1] What are the latest feature cleaning approaches in DC\&ML?
    \item[RQ1.2] What are recent advances in label cleaning in DC\&ML?
    \item[RQ1.3] What are the novel developments in entity matching in DC\&ML?
    \item[RQ1.4] What are the most recent innovations in outlier detection in DC\&ML?
    \item[RQ1.5] What are the latest imputation approaches in DC\&ML?
    \item[RQ1.6] What are recent advances in holistic data cleaning in DC\&ML?
\end{itemize}

\section{Statistical results}\label{sec:results:stats}

\begin{figure}[t]
 \centering
 \includegraphics[width=0.7\textwidth]{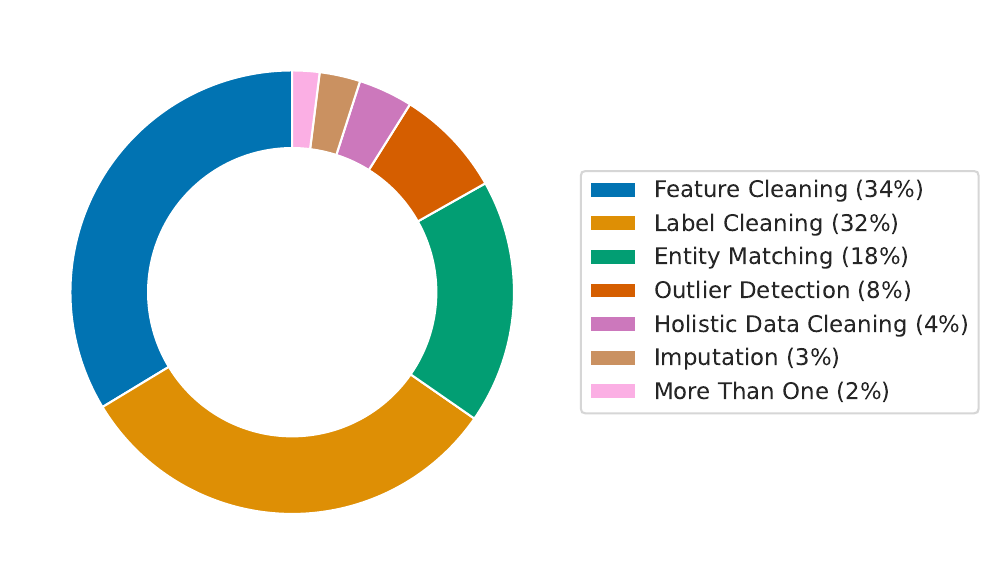}
 \caption{Distribution of data cleaning topics}
 \label{fig:papers_per_task}
\end{figure}

\begin{figure}
     \centering    
     \begin{subfigure}[t]{0.45\textwidth}
         \centering
         \includegraphics[width=\textwidth]{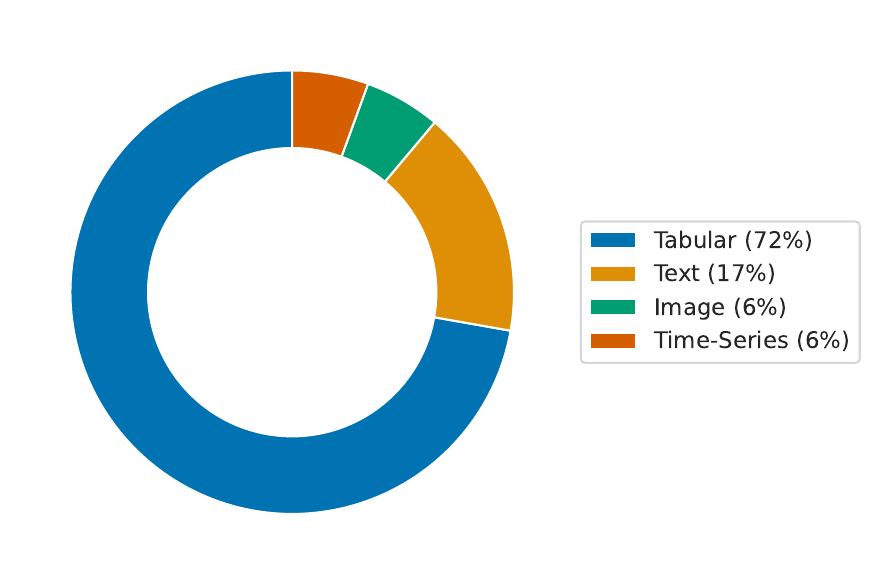}
         \caption{Feature cleaning}
         \label{fc:datatype}
     \end{subfigure}
     \hfill
     \begin{subfigure}[t]{0.45\textwidth}
         \centering
         \includegraphics[width=\textwidth]{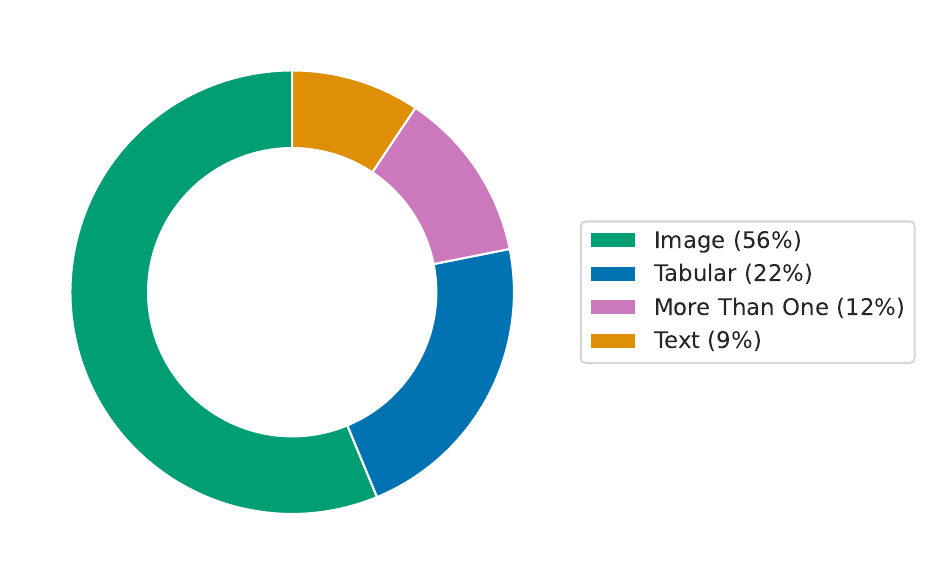}
         \caption{Label cleaning}
         \label{lc:datatype}
     \end{subfigure}
     \begin{subfigure}[t]{0.45\textwidth}
         \centering
         \includegraphics[width=\textwidth]{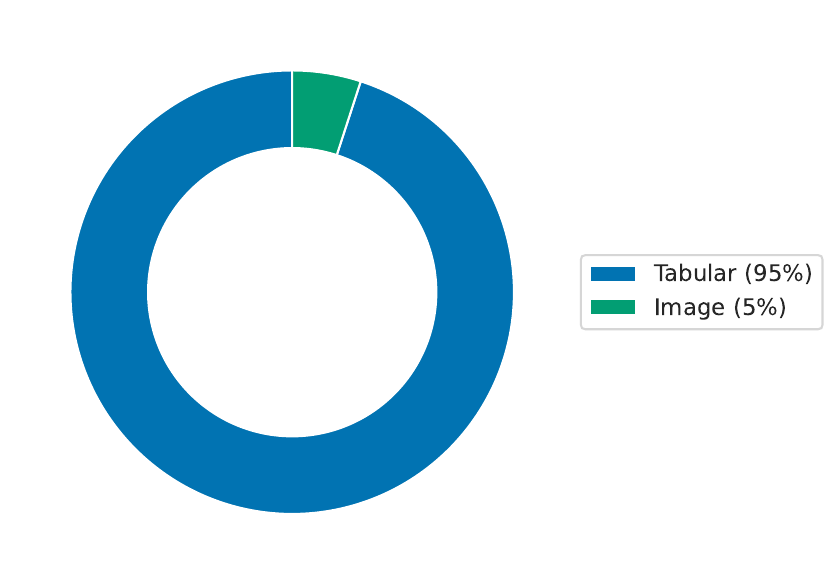}
         \caption{Entity matching}
         \label{em:datatype}
     \end{subfigure}
        \caption{Data types processed per data cleaning task}\label{sr:datatypes}
\end{figure}

\begin{figure}[t]
 \centering
 \includegraphics[width=0.5\textwidth]{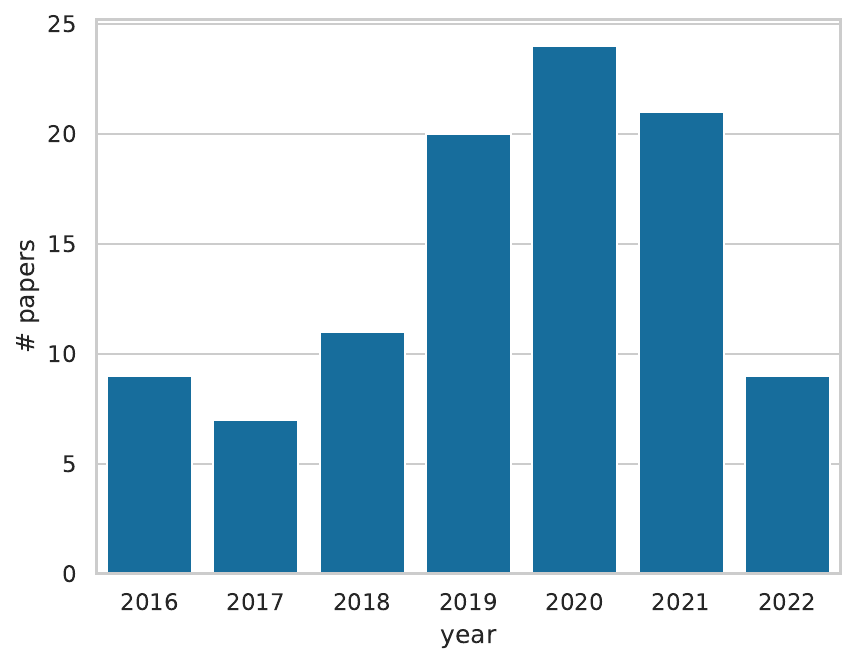}
 \caption{Number of papers published per year}
 \label{fig:papers_per_year}
\end{figure}

\begin{figure}[t]
 \centering
 \includegraphics[width=0.5\textwidth]{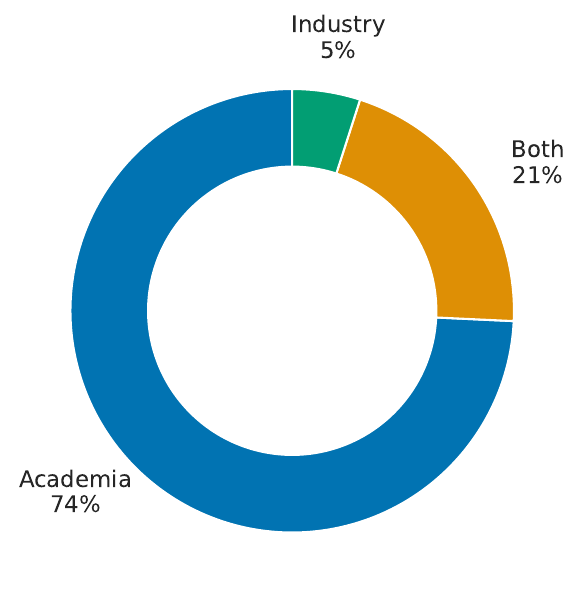}
 \caption{Authors’ Affiliation Distribution}
 \label{fig:authors_affiliation}
\end{figure}

\begin{figure}[t]
 \centering
 \includegraphics[width=0.6\textwidth]{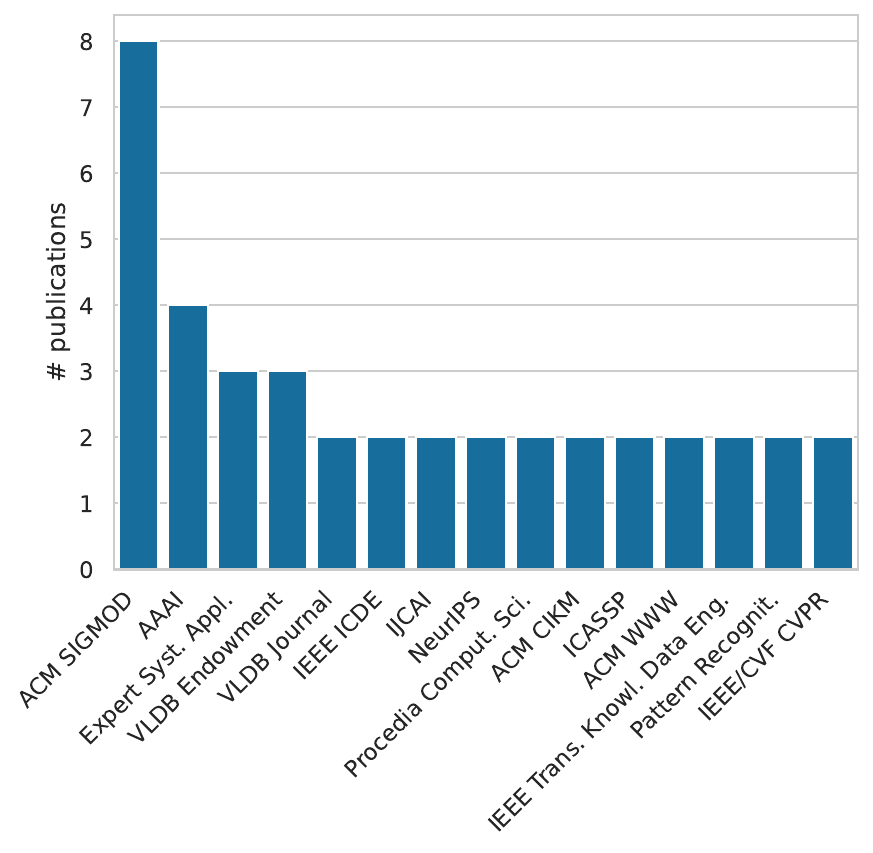}
 \caption{Number of papers published per venue}
 \label{fig:papers_per_venue}
\end{figure}

In this section, we provide a statistical description of the papers included in this study. Figure \ref{fig:papers_per_task} displays the distribution of papers across data cleaning activities (which we previously defined in Section \ref{sec:results:tax}). Studies proposing an approach to clean more than one error at a time are in the category ``More than one''. Generally, the papers that were included in that category proposed an approach that could be adapted for different cleaning goals (e.g., \citep{wang2022sudowoodo}). 
We can observe three main categories of papers, along with four minor ones. The main categories are ``Feature Cleaning'' (34\%), ``Label Cleaning" (32\%), and ``Entity Matching'' (18\%). The minor categories are ``Outlier Detection'' (8\%), ``Holistic'' (4\%), ``Imputation'' (3\%), and ``More than one'' (i.e., approaches that individually fix more than one type of error) (2\%). We display in Figure \ref{sr:datatypes} the data type processed in the experiments of an approach for various data cleaning activities. We plotted this data for the top three data cleaning activities with the highest number of papers, i.e. feature cleaning, label cleaning, and entity matching. We can observe that the data type that is the most commonly processed throughout all data cleaning activities is tabular data. As can be seen in Figure \ref{lc:datatype}, label cleaning is the activity with the highest diversity in terms of data types. In stark contrast, entity matching approaches are almost exclusively processing tabular data, as displayed in Figure \ref{em:datatype}. Figure \ref{fig:papers_per_year} shows the distribution of the papers' publication year. We can observe a generally growing trend. The number of papers in 2022 is less than in 2021 because we collected the papers during this year, in October 2022. In Figure \ref{fig:authors_affiliation}, we display the authors' affiliation distribution. The majority of the papers included in our study have been written by researchers from academia (74\%). We observe that 21\% of the papers included in our study have been written by teams composed of a mix of researchers from academia and the industry. Finally, only 5\% of the papers are attributed to teams from the industry. In Figure \ref{fig:papers_per_venue}, we display the number of papers published per venue. We filtered out venues that counted only one paper for readability purposes. The most recurrent venues are the ACM SIGMOD conference, the Association for the Advancement of Artificial Intelligence conference (AAAI), the Expert Systems with Applications journal, and the VLDB Endowment conference.

\section{Review}\label{sec:review}
In this section, we summarize current data cleaning approaches for tabular, text, and image data. We use the taxonomy described in Section \ref{sec:results:tax} to structure our review. Hence, the following subsections discuss about feature cleaning, label cleaning, entity matching, outlier detection, imputation, and holistic data cleaning. In addition to summarizing the literature, we also provide an analysis of the metrics used for evaluation along with a comparison of the different approaches. We provide a brief description of the existing approaches along with future work recommendations at the end of every subsection.

\subsection{Feature Cleaning}\label{subsection:fc}
\begin{figure}[t]
 \centering
\includegraphics[width=0.85\textwidth]{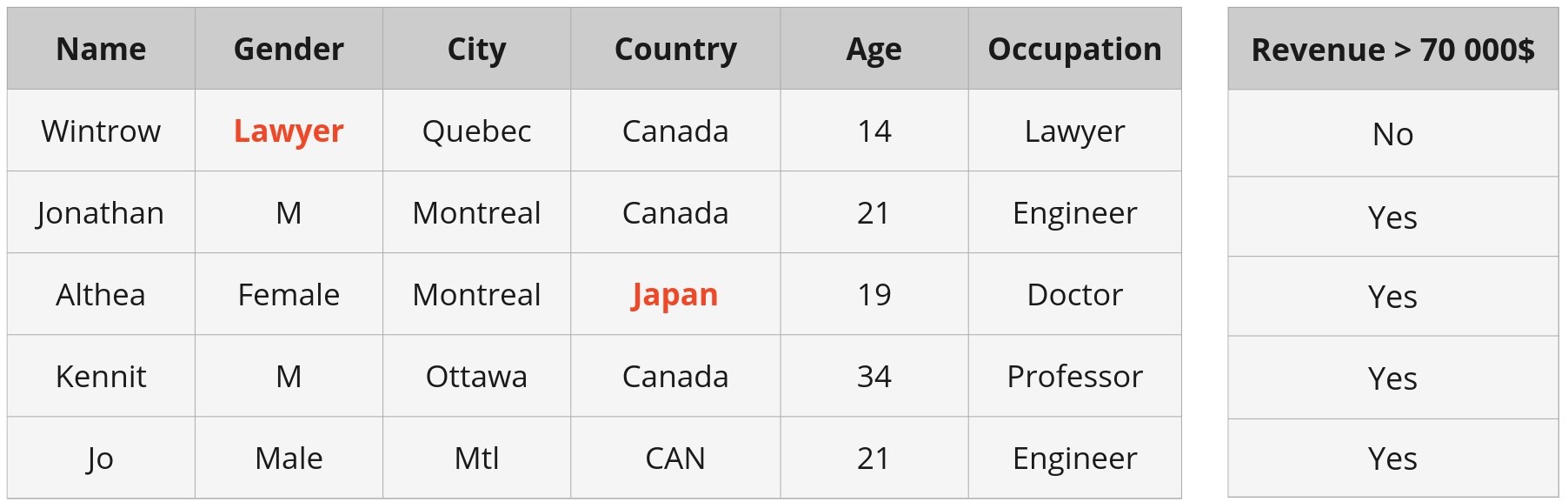}
 \caption{A table with feature errors (highlighted in red).}
 \label{fig:fc_overview}
\end{figure}

In this section, we cover data cleaning approaches to detect or repair feature errors. To better illustrate the type of error cleaned by feature cleaning approaches, we provide a table with feature errors in red in Figure \ref{fig:fc_overview}. Because of the long-lasting efforts of the database community to clean tabular data \citep{rahm2000data, fox1999maintaining, mayfield2010eracer}, the majority of the papers in this section process this data type. Thus, the following subsections present feature cleaning approaches for tabular data, except for the last one (Section \ref{fc:other_data_types}). In the first one (Section \ref{fc:model-based}), we present model-based approaches. Then, we follow with ensemble-based (Section \ref{fc:ensembling-approaches}), transformer-based (Section \ref{fc:transformer}), and autoencoder-based approaches (Section \ref{fc:autoencoder}), which all are subcategories of model-based approaches. We continue with error prioritization (Section \ref{fc:error-prioritization}) and data-cleaning rule generation approaches (Section \ref{fc:dc-rule-gen}). Then, we present other feature-cleaning approaches that do not belong to any of the aforementioned groups (Section \ref{fc:others}). We continue with a description of approaches commonly used to improve the data cleaning approaches performance (Section \ref{fc:improving-dataset}) and we finish with feature cleaning approaches for other data types than tabular (Section \ref{fc:other_data_types}). Finally, we describe how feature cleaning tasks are evaluated in Section \ref{fc:metrics} and provide a comparison of the feature cleaning approaches in Section \ref{fc:comparison}.

\subsubsection{Model-Based Approaches} \label{fc:model-based}
The approaches presented in this category view data cleaning as a prediction problem. That is, a model must predict whether a value is clean or not (i.e., error detection) or what is the clean value of a dirty cell (i.e., error repair). Formally, for error detection, a model $m$ must predict whether a record's feature value $r[A_i]$ (where $r$ refers to a record and $A_i$ to the i-th attribute) is dirty or not ($y \in [0,1]$) given the full record $r$. Similarly, for error repair, a model $m$ must predict the clean value $y$ for a record's feature $r[A_i]$ given the full record $r$, where $y$'s domain is not necessarily restricted. Note that the following sections (i.e., ensemble-based, transformer-based, and autoencoder-based approaches) are also model-based approaches. In the following, when referring to records' labels, we refer to the label used by a data cleaning approach to learning to clean data.

One of the major differentiating factors between error detection approaches is their choice of feature engineering. We discerned three categories of features: (1) frequency-based features, (2) format-based features, and (3) metadata features. Similar to outlier detection approaches (see Section \ref{sec:od}), the engineered features are generally designed to highlight the abnormal characteristics of data. A commonly used feature-engineering practice for error detection is to create features that measure how frequent a value is in a dataset (i.e., frequency-based features) \citep{visengeriyeva2018metadata, neutatz2019ed2, heidari2019holodetect, pham2021spade}. For example, the authors in \citep{neutatz2019ed2} measured the TF-IDF score of n-grams inside a cell to encode how common a cell's value is. The less common it is, the higher the chances of it being an error. Frequency-based features can be measured using the information contained in other columns \citep{neutatz2019ed2, heidari2019holodetect}. For example, \citet{neutatz2019ed2} used the co-occurrence of values among different attributes to facilitate the error detection process. Another category of engineered features focuses on the format of the values (i.e., format-based features) \citep{visengeriyeva2018metadata, neutatz2019ed2, heidari2019holodetect, pham2021spade}. Corrupted cells might not follow the syntactic format that is expected for a feature. For example, a value that represents a name should not have numbers. \citet{pham2021spade} replaces all numbers and characters with symbols that are unique to them; thus focusing only on the shape of the data. For example, given the value ``400\$'', the encoded format could be ``nnns'', where ``n'' represents numbers and ``s'', symbols. Finally, the data cleaning approaches sometimes complement their feature set with metadata information (i.e., metadata features) \citep{visengeriyeva2018metadata, neutatz2019ed2}. For example, \citet{neutatz2019ed2} indicates the data type and the string length of a cell in the feature set.
For error repair, ML can be used to predict values to repair corrupted cells, similar to data imputation (see Section \ref{imputation}), as it was done in \citet{ataeyan2020novel}.

\subsubsection{Ensemble-Based Approaches} \label{fc:ensembling-approaches}
Coined by \citet{neutatz2021cleaning}, ensemble-based approaches to data cleaning comprise every approach where a model uses the output of data cleaning tools to clean data. As it was pointed out in \citet{abedjan2016detecting}, no single data cleaning tool is the best in every situation and no tool can detect all errors. Leveraging ensembles of data cleaning tools opens up the possibility of powerful approaches to clean errors since they could potentially detect and repair a larger diversity of errors \citep{rekatsinas2017holoclean, visengeriyeva2016improving, mahdavi2019raha}. Similarly to ensembling techniques with ML models, ensemble-based approaches for feature cleaning have an ensemble of data cleaning tools, which we refer to as the ``base'' data cleaning tools. The outputs of the base data cleaning tools are ingested by one singular model making the final data cleaning prediction, which we refer to as the meta-model.

Usually, the base data cleaning tools are simple cleaning tools developed without using ML techniques. The most commonly used ones are integrity constraints \citep{rekatsinas2017holoclean, mahdavi2019raha, visengeriyeva2018metadata, visengeriyeva2016improving}, matching dependencies \citep{rekatsinas2017holoclean, mahdavi2019raha}, outlier detection \citep{rekatsinas2017holoclean, mahdavi2019raha, visengeriyeva2018metadata}, and pattern violation detection \citep{mahdavi2019raha, visengeriyeva2018metadata}. For data repair, simple algorithms that memorize transformations between dirty and clean data are sometimes used \citep{mahdavi2020baran, wang2022sudowoodo}. These are designed specifically for ensemble-based approaches and can only suggest good repairs collectively. Hence, they are generally not viable data repair tools independently. Mahdavi et al. define three types of such tools: (1) value-based models, (2) vicinity-based models, and (3) domain-based models \citep{mahdavi2020baran}. Models in the first category only consider the corrupted value when suggesting repairs \citep{mahdavi2020baran, wang2022sudowoodo}. For example, given a misspelled string such as ``a dg'', the tool could suggest ``a dog''. On the contrary, vicinity-based models leverage the context of the corrupted value to predict its repair. For example, given the ``city'' attribute of a record is set to ``Tokyo'', then we know that the ``country'' attribute should be ``Japan''. Finally, domain-based models repair a value based on the domain of the attribute. For example, a domain-based repair strategy could replace a corrupted value with the attribute's mode.

Once the base data cleaning tools have processed an example, one is left with a set of data cleaning predictions (error detection or repair) that are ingested by the meta-model for the final data cleaning prediction. We observed different ways to include the cleaning signal of the base data cleaning tools in the prediction process of the meta-model. When the meta-model of the cleaning approach is a Probabilistic Graphical Model (PGM), the base data cleaning tools are transformed into inference rules for the meta-model \citep{rekatsinas2017holoclean, visengeriyeva2016improving}. For error detection, \citet{mahdavi2019raha} trains, for each feature, a model that ingests the base data cleaning tools' binary output (i.e., dirty or not) and predicts whether a record is dirty or not. Compared to error detection, error repair with ensemble-based approaches has the additional challenge of having base data cleaning tools that may output a wide variety of values. Thus, error repair tools adopt a different approach to combine the base data cleaning tool predictions. Instead of generating a repair, the meta-model selects the repair that is the most likely to be correct in a pool of repairs suggested by the base data cleaning tools \citep{mahdavi2020baran, wang2022sudowoodo}. In \citet{mahdavi2020baran}, each base error repair tool suggests values to repair corrupted cells. Then, for each repair, a vector holding the confidence that the repair is correct according to each base error repairer is generated. This vector is then fed to a Neural Network (NN) that will predict how likely a repair is given the confidence of the base error repair tools. The repair with the highest score becomes the repaired version of a dirty record. Instead of evaluating repairs using the confidence score of the base tools, Wang et al. let the meta-model select the most probable repair by comparing the repaired versions of a record with its dirty version \citep{wang2022sudowoodo}. To facilitate the comparison process, the records are transformed into embeddings. Then, an NN ingests the difference between the embeddings, along with the embedding resulting from the concatenation of both records. This strategy of encoding records captures both the differences and similarities between the compared records. For every considered pair, the NN will output a score, stating how likely is the proposed repair. The most likely repair for a record is then selected. This approach was originally designed for entity matching but showed to be apt for error repair as well. 

While ensemble-based approaches may seem appealing, they have a few challenges of their own. To begin with, the predictions of a meta-model can be skewed if the predictions of the base tools are correlated \citep{visengeriyeva2018metadata}. This is simply due to the fact that a larger feature space is allocated to what is essentially the same information. The authors in \citep{visengeriyeva2018metadata} address this challenge by pruning out correlated tools and keeping the most effective ones. To achieve that, k-means clustering is run on the predictions of the base tools. Then, using a validation dataset, the best tool for each cluster is selected. In addition to the aforementioned issue, a large pool of base cleaning tools results in two additional problems. First, it may lead to a longer run time for the whole data cleaning process. Thus, \citet{mahdavi2021semiRahaBaran} prunes out tools that proved to be ineffective on similar datasets. The dataset similarity is measured by comparing the semantic and syntactic similarity between columns. Second, significant human labor can ensue if a large number of tools require manual parametrization. To address this problem, \citet{mahdavi2021semiRahaBaran} generates, for every tool, all configurations in a predefined set.

\subsubsection{Transformer-Based Approaches} \label{fc:transformer}
In this section, we present approaches that clean the features of a record using transformer models \cite{vaswani2017attention}. Because of their impressive performances in the last years, there has been an interest in the research community to apply this kind of model to a wide variety of tasks \citep{chen2021evaluating, dosovitskiy2020image, chasmai2021cubetr}. Two issues must be circumvented in order for these models to be used for feature cleaning. The first hurdle is in the very nature of the task: feature cleaning is different from text generation. Effectively, transformer models are designed to generate text, while feature cleaning is usually conceptualized as a classification problem. Three approaches have been suggested to address this problem. The first solution transforms feature cleaning into a text generation problem. More precisely, \citet{narayan2022can} prompts the foundation models with data-cleaning questions. For example, for error detection, the user may ask the model if there is an error in a given sample, to which the model will answer ``yes'' or ``no''. The authors further show that very few training samples are required for the model to detect errors, from a few (few-shot learning) to none (zero-shot learning). For error repair, \citet{tang2020relational} uses a transformer to predict the value of every cell of a record and replace it if it differs from the shown value. The transformer model is trained using the masked data model objective. That is, the model ingests a record with a masked value and outputs a record with the mask replaced with a value. If the value generated by the model differs from the actual value that was masked, then the example is considered to be corrupted. Instead of transforming data cleaning into a text generation task, \citet{nashaat2021tabreformer} adapts a transformer model to a classification task. A linear layer with softmax activation is added to the output of the transformer and the model is trained to minimize cross-entropy loss. Before being fine-tuned for this binary classification task, the transformer is trained using the masked data model objective, so it learns the structure of data. The third way to use transformers for data cleaning is by using them to generate embeddings that can be used by another model for data cleaning. \citet{wang2022sudowoodo} pre-trained a transformer using SimCLR contrastive learning framework \citep{chen2020simple}, so similar records have similar embeddings. Then, an NN ingests these embeddings and predicts how likely a repair is given a dirty sample. The most likely repair is selected. 

The second issue that must be addressed before using transformers for data cleaning is the format of the data itself. While the data cleaning approaches in this section are applied to tabular data, transformers are trained with natural language. A naive approach would feed the record as it is to the model (i.e., transform the record into a sentence by concatenating its features using spaces). However, precious information about the features' semantics would be lost, since the model does not know what a value represents (e.g., a number does not make sense if we do not know what it counts). Hence, to address that problem, the tabular records must first be transformed into sentences, and then the model must learn to understand the structure of these sentences. To inject the tabular structure of data into strings, \citet{narayan2022can} prepends each attribute value with its name before concatenating it into a string. For a record that has a feature named ``City'' and a value of ``Montreal'', then the attribute would be represented by ``City: Montreal''. The latter would be concatenated into a sentence with the substrings generated for the other features. To inject the semantic difference between attributes' names and their values, \citet{tang2020relational, wang2022sudowoodo} further prepend attributes' names with the special token ``[A]'' and the values with ``[V]''. Extending on the previous example, we would have ``[A]City [V]Montreal''. Once tabular data is converted into strings, the model must learn to understand the format of the data. Similar to what is usually done with large language models (LLM), the masked data model objective is used in \citet{nashaat2021tabreformer, tang2020relational} to train transformers in an unsupervised way. The model is trained to predict the missing value in a record's entry, which enables it to unlearn its assumption of processing natural language data and learn the structure of the dataset at hand. \citet{wang2022sudowoodo} leveraged contrastive learning to train the encoder of a transformer to generate embeddings that are close in the latent space for similar records, and far for different records.

\subsubsection{Autoencoder-Based Approaches} \label{fc:autoencoder}
The data cleaning approaches in this section use autoencoders to detect or repair records. Autoencoders are a type of ML model trained to output the data they receive as input while being constrained to represent the data in a latent space with fewer dimensions than the input \citep{bank2020autoencoders}.

\citet{mauritz2021probabilistic} argues that autoencoders have denoising capabilities because of the compression process. Representing the input data into a lower dimension forces the autoencoder's intermediate representations to contain only the most important information necessary to recreate the input data. As a result, noise is removed from the data. Hence, they use the innate noise removal capability of autoencoders to repair records. To do so, they first transform records into probabilistic ones (i.e., records whose features have a probability distribution over potential values). Then, they feed the record to the autoencoder, which outputs another probabilistic record. The result can be transformed back to non-probabilistic data by picking the most likely value for each feature. \citet{tonolini2020tomographic} proposes a variant of variational autoencoders (VAE) architecture for noise removal that does not suffer from the posterior collapse problem (when the model confidently suggests a small number of reconstructions). Authors employ a three-step process to clean samples using their autoencoder. First, an example to clean is ingested by the model and a latent distribution is generated by the model. Second, multiple latent vectors are sampled from that latent distribution. Third, examples are generated from each latent vector and averaged to generate the final clean value. Instead of using the autoencoder to generate values to repair errors, Liu et al. use the reconstruction loss of an autoencoder to detect corrupted records \citet{liu2022picket}. Records that have a high loss during the first epochs of training are flagged as erroneous. The study further shows that poisoned samples generally have the lowest loss; hence their method can also be used to detect such records.

\subsubsection{Error Prioritization Approaches} \label{fc:error-prioritization}
Instead of directly cleaning data, the approaches covered in this section try to identify and prioritize the records that should be cleaned by a human expert. 

Authors in \citet{zhang2018deepclean} leverage the knowledge stored in Wikipedia to clean data. More precisely, by generating questions for a QA engine that understands natural language and whose answers are Wikipedia pages, they are able to verify the values in a dataset by comparing them with what is on Wikipedia. If a value conflicts with what the engine has returned, the records are sent to a human for review. Instead of identifying records that are potentially erroneous, authors in \citet{karlavs2020nearest} search for records that would have the largest impact on a model's performance if cleaned. They argue that data cleaning efforts are wastefully spent on samples that do not alter a model's performance. Thus, the authors propose an approach that tries to find samples that, once cleaned, will render the next iterations of cleaning less useful (since the next iterations will not alter a model's performance). Similarly, Krishnan et al. \citep{krishnan2016activeclean} propose an approach to prioritize the samples sent to a human for review. A model is first trained on a dataset (for the prediction task of the dataset, not for data cleaning). Then, samples that would trigger the largest weight update if the model was retrained on them after being cleaned are prioritized. The weight updates are estimated using previous examples with similar feature values.

\subsubsection{Data Cleaning Rule Generation Approaches}\label{fc:dc-rule-gen}
The approaches included in this category use ML to infer data cleaning rules that can be used to clean a dataset. Data cleaning rules are typically expressed using integrity constraints, a technique from database schema design \citep{abiteboul1995foundations}. Integrity constraints can express a wide variety of logical clauses, such as: ``two persons living in the same city must live in the same country'' or ``two persons can not have the same phone number'' \citep{ilyas2019data}.

\citet{ge2020hybrid} proposes a hybrid data cleaning framework, named MLNClean, that leverages Markov Logic Networks (MLNs) to supplement integrity constraints and makes it possible to clean unknown data errors. MLNClean first uses MLN to generate a set of probable data cleaning rules with their corresponding probabilities. These rules are then leveraged to generate repair candidates. The repair candidate that is the most likely (based on the likeliness of the data cleaning rules that generated it) and that differs the least from the original value is then used to repair an observed value. In another work \citep{guo2019learning}, authors propose AutoFD, a framework for Functional Dependency (FD) (a type of integrity constraint) discovery based on structure learning techniques in PGMs \citep{guo2019learning}. A PGM is introduced to capture dependencies that FDs introduce among attributes in a dataset, revealing that learning the graph structure of this model is equivalent to discovering FDs. 

\subsubsection{Other Approaches}\label{fc:others}
In this section, we cover other feature cleaning approaches for tabular data that are not included in the previous categories.

\begin{itemize}
    \item \citep{krishnan2017boostclean}: Similar to ensemble-based feature cleaning, their work uses an ensemble of feature cleaning tools. However, contrary to the aforementioned category, their goal is not to clean data but rather to directly generate a model that would have been trained on clean data. To do so, all feature cleaning tools are individually applied to a dataset, generating different versions of it. Then an ML algorithm trains a model for each processed dataset. The models' predictions are combined together using boosting \citep{wikipedia_2023_boosting}. As a result, the predictions of the ensemble are influenced by all the data cleaning tools.
    \item \citep{lew2021pclean}: Authors propose a new domain-specific Probabilistic Programming Language (PPL), namely PClean that is tailored for data cleaning. Similar to \citet{rekatsinas2017holoclean}, PClean leverages dataset-specific knowledge. Authors introduce a domain-general nonparametric prior on the number of latent objects and their link structure in the dataset. Code written by PClean customizes this prior probability via a relational schema (for relational dataset) and via generative models for objects’ attributes. Inference in PClean is based on a novel Sequential Monte Carlo (SMC) algorithm, to initialize a latent object database with plausible guesses, and novel restoring updates to fix mistakes.
   
    \item \citep{oliveira2020batchwise}: Their paper proposes an improvement over HoloClean \citep{rekatsinas2017holoclean}. Most of the data cleaning approaches consider that data cleaning is only applied once, before training a model. This does not reflect how data is cleaned with real ML software systems since new training data is continuously generated. Hence, they propose a few enhancements to HoloClean \citep{rekatsinas2017holoclean} to make it more efficient in that context. Notably, they avoid retraining the data cleaning model if the distribution of incoming data does not change.
\end{itemize}

\subsubsection{Common Techniques to Improve Performance} \label{fc:improving-dataset}
In this section, we cover various techniques commonly used to improve a model's performance for model-based approaches, namely, data augmentation, semi-supervised techniques, and active learning.

\paragraph{Data Augmentation}
Data augmentation is a technique to generate synthetic new examples by applying transformations to existing data \citep{hernandez2018data}. Data augmentation is used in feature cleaning approaches to address the class imbalance problem; training datasets for feature cleaning have a larger amount of labeled clean records than labeled dirty records \citep{nashaat2021tabreformer, heidari2019holodetect}. In other words, in labeled datasets, there is generally more clean data than dirty data and dirty data is needed for the ML approach to learn to detect dirty data (error detection) or to learn to clean it (error repair). Hence, data augmentation tries to address that problem by generating more labeled dirty data. To do so, data augmentation techniques transform clean data into dirty data by applying plausible corruption mechanisms (i.e., the error induced into the record is plausible given the dataset). In order to define a data augmentation strategy, two components must be defined: the process that transforms clean data into dirty data and the process that selects data transformations (i.e.,  assigning a probability to each transformation).
Instead of manually crafting data transformation rules, \citet{nashaat2021tabreformer, pham2021spade, heidari2019holodetect} learn them from the data. In \citet{nashaat2021tabreformer}, Gestalt Pattern Matching \citep{tawfik2020evaluating} is used to find the non-matching substrings between the clean and dirty versions of a text feature. These substrings are then used to form a new mapping, from the clean substring to the dirty one. Similarly, \citet{pham2021spade, heidari2019holodetect} adopt a hierarchical pattern matching approach to learn valid transformations.   
Once data transformation functions are defined, the data augmentation approach must build the function that selects the data transformation that will be applied to a record. In other research works \citet{nashaat2021tabreformer, pham2021spade, heidari2019holodetect}, authors sample a valid transformation following the empirical distribution of transformations (i.e., how often each transformation was used empirically).

\paragraph{Semi-Supervised Techniques} 
Semi-supervised learning is a branch of ML that uses unlabeled data (in addition to labeled data) to build better models \citep{pise2008survey}. The semi-supervised techniques presented in this section use the labeled data to label the unlabeled data. In order to generate training data for its data augmentation approach, \citet{heidari2019holodetect} fits a Naive-Bayes model that, similar to imputation, predicts a clean value for each feature of a record. The most confident predictions (i.e., with over 90\% confidence) become the clean versions of the observed values. \citet{pham2021spade, mahdavi2019raha} use label propagation to increase a labeled dataset size for error detection. Label propagation consists of assigning the label of a record to a similar unlabeled record \citet{zhur2002learning}. The challenge with this approach is to devise an accurate measure of similarity between records to avoid mislabels. \citet{pham2021spade} considered two feature vectors to be similar if all their attributes did not differ by more than a threshold. Similarly, \citet{mahdavi2019raha} used hierarchical clustering \citep{aggarwal2013data} to find groups of similar feature vectors. Records in the same cluster share the same label.

\paragraph{Active Learning}
Active learning refers to a situation where a ML algorithm can iteratively query an oracle (e.g., a human) to label new data points \citep{wikipedia_2023_active_learning}. In the following, we show how some studies used active learning to achieve superior performance. \citet{neutatz2019ed2} relies on the confidence score of the data cleaning models to select the cells to send to a reviewer. Because they ask reviewers to clean cells, not tuples, their acquisition function (i.e., the function that selects records) starts by selecting a feature to clean (i.e., a column of a dataset), then picks a batch of cells inside that column to clean. The column selected in the first step is the one where the model is, on average, less confident of its predictions. The cells are then selected using the query-by-committee algorithm \citep{freund1997selective}. \citet{pham2021spade} uses two different models for data cleaning and active learning. The model used for active learning is implemented using Probabilistic Soft Logic \citep{bach2017hinge} and, similar to the data cleaning model, tries to predict whether cells are dirty or clean. Records that are more likely to be dirty are sent for review. Finally, for error repair, \citet{mahdavi2020baran} prioritizes tuples with a lot of errors that are common in the dataset. The goal is to create a training dataset that covers most types of errors in a dataset. Note that the approach assumes that an error-detection tool already marked the dirty cells.

\subsubsection{Other Data Types} \label{fc:other_data_types}
In this section, we cover feature cleaning approaches for other data types than tabular. We chose to present data cleaning approaches for each data type in a dedicated part because each data type has a peculiarity that makes the data cleaning task significantly different. Thus, in the following, we describe feature cleaning approaches for time series, text, and image data, respectively.

\paragraph{Time Series}
Time series are a collection of observations made in chronological order \citep{wang2019time}. The feature cleaning approaches in this category have to consider the ordinal nature of time series data in order to clean it. 

\citet{zhang2020time} combined auto-regressive models with the minimum repairing principle of data cleaning \citep{ilyas2019data} to iteratively clean data. Auto-regressive models are used to suggest values to repair anomalous observations. In their approach, an auto-regressive model suggests repairs for every observation. Only one repair is applied: the one that differs the less from the original observation. The auto-regressive model is retrained on the cleaned time series, and the procedure is repeated until the difference between the original and clean value is below a threshold or the maximum number of iterations is reached. Similarly, \citet{akouemo2017data} presents an iterative approach for the detection and repair of anomalies in time series data. First, Auto-regressive with Exogenous inputs (ARX) and NN models are trained on a time series and used to predict repair values for every value of a time series. Then, hypothesis testing over the difference between observed values and predicted values is used to detect outliers. The most abnormal observed value is replaced by the predicted value, and the process is repeated. 

\paragraph{Text}
In this section, we cover techniques for detecting errors in text data. The addressed errors include grammatical, syntactical, and semantic errors. 

NNs models are used for detecting and correcting grammatical errors in text. For example, in \citet{he2021automatic}, an RNN-based model is used for detecting grammatical errors in English verbs. The model consists of two sub-RNNs such that one processes the sentence from the beginning to the target verb and the other processes the sentence from the end to the target verb. An additional layer is used to process the output of the RNNs and predict the verb which is then compared with the original verb to detect any potential error. Knill et al. approach error detection as a sequence labeling task where a Bi-directional LSTM is trained to classify each token in the input sentence as correct or incorrect within the sequence \citet{knill2019automatic}. \citet{wang2020grammatical} utilize word contextual embeddings generated by BERT and concatenate them to character embeddings generated by an LSTM. They then use the BERT model, in particular, the first six layers of BERT to classify each word as correct or incorrect. \cite{rei-yannakoudakis-2016-compositional} investigates different NN architectures (CNN, Bi-RNN, Bi-LSTM, and their multi-layer variants). Provided with a sequence of tokens, the model calculates the probability for each token to be correct or incorrect. Among the compared models, Bi-LSTM outperforms all other models including the multi-layer Bi-LSTM variant.\\
To detect syntactic errors, \citet{santos2017finding} proposes to train two LSTM models, one is forward and the other is backward, to predict tokens. The disagreement between the two models is used as an indication of a possible syntax error. They also attempted to propose a correction by identifying an alternative token sequence derived from the models.\\
The model introduced by \citet{spithourakis2016numerically} detects semantic errors in terms of inconsistencies between a numerical value (e.g., clinical test result) and its interpretation. When a document is received, sets of words with similar meanings are employed to generate all possible word substitutions within the document, creating a set of candidate repaired documents. Each of the candidate documents is fed to an LM to predict which one is the most likely to be correct. Likelihood ratios between candidate and original documents are then computed. Candidates with higher scores than the original indicate errors; the highest score is accepted as the correction.

\paragraph{Image}
A common distortion that can be encountered in images, particularly in those captured by autonomous vehicle perception systems is a radial blur. This type of blur arises due to the motion of the imaging system during image acquisition. \citet{hurakadli2019deep} proposes a deep-learning-based pipeline to estimate and correct the radial blur in order to improve the detection of traffic signboards in images. The proposed pipeline consists of estimation and enhancement modules. The authors designed a convolutional neural network (CNN) named CuratorNet, to estimate the point spread function (PSF) of an image with a precision of up to a second decimal point. PSF is a mathematical description of how light gets spread out or blurred across the image due to various factors in the imaging system. By applying the inverse operation of the PSF to the observed image, it can be possible to recover the original scene. To correct the image, the authors propose a convolutional autoencoder-based enhancement module that eliminates the radial blur, based on the obtained PSF estimation.

\subsubsection{Evaluation metrics} \label{fc:metrics}
Error detection and repair approaches are often evaluated like classification techniques. Hence, evaluation metrics used in the field of ML, such as accuracy, F1 score, precision, and recall are most commonly used \citep{wang2022sudowoodo, nashaat2021tabreformer, pham2021spade, neutatz2019ed2, heidari2019holodetect}. For image and time series data, different metrics are utilized. Peak-signal-to-noise is used in image cleaning research to measure the amount of noise removed from a picture \citep{tonolini2020tomographic, hurakadli2019deep}. Regression metrics, such as mean square error, are used in time-series to evaluate how close a cleaned sample is to its correct value \citep{zhang2020time}. Since most approaches rely on a set of labeled samples to learn to detect or repair data, a few compare the performance of their approach with varying amounts of labeled data \citep{neutatz2019ed2, karlavs2020nearest, nashaat2021tabreformer}. For example, the authors in \citet{neutatz2019ed2} evaluate the performance of their approach trained on datasets ranging between 20 samples to 300. By following this strategy, they can measure how effectively their approach learns to clean data. Instead of evaluating the correctness of the predictions of the data cleaning approach, a subset of studies measured the impact of data cleaning on a downstream ML model \citep{krishnan2016activeclean, liu2022picket, krishnan2017boostclean}. Other than correctness, the execution time of the approach is sometimes measured \citep{mahdavi2019raha, liu2022picket, neutatz2019ed2, rekatsinas2017holoclean}, since it is sometimes significant. For example, the approach proposed in \citet{rekatsinas2017holoclean} takes over 6 hours to process a 2M-examples dataset.

\subsubsection{Comparison of the Approaches} \label{fc:comparison}
Among the various types of feature cleaning approaches presented, one category stands out with its performance: transformer-based approaches. They tend to outperform approaches from other categories on a large number of datasets, for error detection and repair \citep{wang2022sudowoodo, narayan2022can, nashaat2021tabreformer}. For error detection, the authors in \citet{narayan2022can} showed a prompting approach that required only 10 samples to achieve better results than  state-of-the-art approaches. Previous model-based approaches achieved comparable scores with only 10 to 20 samples \citep{mahdavi2019raha, mahdavi2020baran}. However, they leveraged data augmentation to make the most out of this limited set of training samples. Previous autoencoder-based approaches detected errors in a dataset without any training samples, albeit with worse performances \citep{liu2022picket}. 
\\
In addition to being performant and requiring a few training samples, transformer approaches require very minimal manual labor to be applied to new datasets. The same prompt formats can be used to process various datasets. In comparison, ensemble-based approaches often depend on data cleaning tools that are dataset-specific. For example, a common cleaning signal ingested by many data cleaning approaches is data cleaning rules (e.g., “a country attribute must be set to Japan if the city is Tokyo”). Thus, to achieve comparable performance to the experiments conducted in the work of \citep{rekatsinas2017holoclean, heidari2019holodetect, mahdavi2019raha}, one must first define data cleaning rules and configure data cleaning tools before cleaning any new dataset. 
\\
While the ensemble-based approaches we reviewed in this paper are generally not as performant as transformer-based approaches, they offer the possibility of integrating the latter approaches in their ensemble of base data cleaning tools and hence do not become obsolete. However, one must be careful integrating too many data cleaning tools into an ensemble-based approach, since it might lead to longer runtimes and burden the user with the configuration and maintenance of many tools, as explained in Section \ref{fc:ensembling-approaches}. 
\\
In general, transformer-based approaches propose solutions that are more performant, consume fewer training samples, and can be easily adapted to new datasets. However, they can be surpassed at times by other approaches. For example, the authors in \citep{pham2021spade} proposed a model-based approach that does not use transformer models and which outperforms transformer-based approaches on some datasets.

\vskip2em
\begin{small}
    
\begin{tcolorbox}[enhanced, breakable, title=Feature Cleaning: Summary and Future Directions]
\noindent\textbf{General existing approaches}: We presented approaches to clean tabular, text, and image data. The feature cleaning approaches for text included in our study fix grammatical, syntactic, and semantic errors using variations of recurrent neural networks or transformer models \citep{DBLP:journals/corr/VaswaniSPUJGKP17}.  The feature cleaning approaches for images included in our study fix image blur using Deep Learning (DL) techniques. We identified the following feature cleaning approaches for tabular data.
\begin{itemize}
    \item \textit{Model-based approaches}: Approaches that train an ML model to directly clean data. One may observe the following three subcategories.
    \begin{itemize}
        \item \textit{Ensemble-based approaches}: An ML model uses the predictions of base data cleaning tools for cleaning. The base data cleaning tools are generally simple.
        \item \textit{Transformer-based approaches}: Approaches that use a transformer to clean the features of a record. They rely on the capabilities of pre-trained LLMs. 
        \item \textit{Autoencoder-based approaches}: Approaches that use autoencoders to clean data. They rely on the denoising capability of auto-encoders. 
    \end{itemize}
    \item \textit{Error prioritization approaches}: Approaches that prioritize the records that should be reviewed by an expert.
    \item \textit{Data cleaning rule generation approaches}: Approaches that generate data cleaning rules and use them to clean data. 
\end{itemize}
\vskip1mm
\noindent\textbf{Future directions}: 
\begin{itemize}
    \item Develop feature cleaning approaches that consider the label of a record when cleaning its features. The approaches covered in our study do not leverage the special status of labels when cleaning a feature (they are not mentioned nor treated differently than other features). Potentially, labels could help clean the features of a record more efficiently than if treated as another feature. Thus, we encourage future works to explore how the records' labels can be used for feature cleaning.
    
    \item Extend the advances done with ensemble-based approaches. Future works should experiment with a wider number of base data cleaning tools. For example, other feature cleaning approaches could themselves be included in the ensemble.
    

    \item Explore how techniques from outlier detection can be transposed to error detection. Outlier detection and error detection are similar challenges since corrupted records often are outliers. Future studies could leverage the existing literature on outlier detection and ML \citep{pang2021deep, nassif2021machine}.

\end{itemize}
\end{tcolorbox}
\end{small}

\subsection{Label Cleaning}\label{label-cleaning}
\begin{figure}[t]
 \centering
\includegraphics[width=0.85\textwidth]{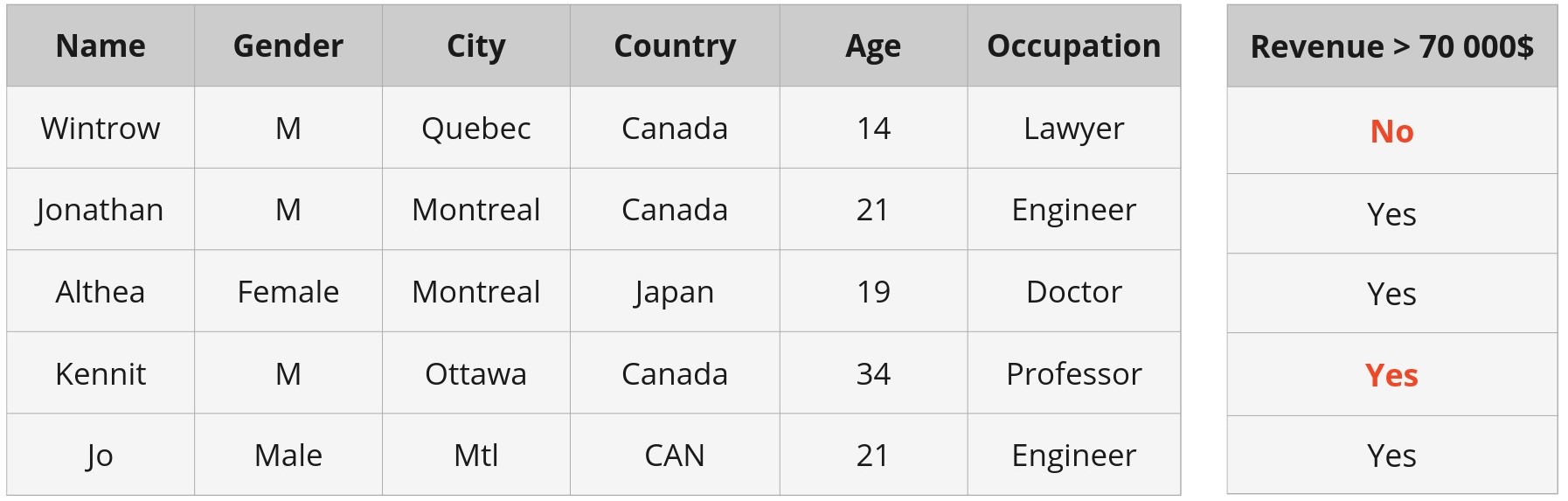}
 \caption{A table with label errors (highlighted in red).}
 \label{fig:lc_overview}
\end{figure}

In this section, we cover data cleaning approaches to detect or correct label errors. To better illustrate the type of error cleaned by label cleaning approaches, we provide a table with label errors in red in Figure \ref{fig:lc_overview}. For simplicity, we refer to label errors as mislabels. We do not differentiate the label cleaning approaches based on type of the data being cleaned, since the main ideas are generic. That is, any approach can be adapted to other data types if an adequate representation learning technique is used.

In the following, we describe the different categories of label cleaning techniques seen in our review. The first category uses the confidence score of an ML model to detect corrupted samples. The second one adopts a similar approach but uses the loss of a model instead of its confidence scores. The third one detects mislabels by searching for instances that degrade a model's performance. The fourth one uses outlier techniques to find mislabels. We also present label cleaning approaches that did not belong to any of the aforementioned categories. We describe how label cleaning tasks are evaluated in Section \ref{lc:metrics} and provide a comparison of the label cleaning approaches in Section \ref{lc:comparison}.

\subsubsection{Uncertainty-Based Approaches}\label{lc:uncertainty}
A straightforward approach to label cleaning is to use the confidence score of an ML model to detect corrupted samples. After having been trained on a dataset, one can suppose that an ML model has learned the representative features of each class. Hence, if the model's predictions disagree with the assigned label for an instance (i.e., the model assigns a low probability for the labeled class), one can assume that the instance has a higher chance of being corrupted compared to other samples. However, the model may predict something different than a sample's label because of corrupted features, mislabels, or more simply because the model made a mistake. Thus, the predictions of the model can be used as a weak signal for detecting mislabels. 

Once a sample is marked as being mislabeled, it must be cleaned. The cleaning strategies sometimes assume that someone will manually clean the flagged instances. When they do, either all the instances are sent for review \citep{rottmann2023automated, teso2021interactive} or only the most important ones. In the latter case, one must find criteria to prioritize samples over others. \citet{rehbein2017detecting} prioritizes records for which the model's predictions have the highest entropy. Similarly, \citet{atkinson2021tsar, atkinson2020identifying} prioritize the records where the model is highly uncertain about the labeled class. In order to avoid uselessly spending humans' time, \citet{bernhardt2022active} proposed to also consider the difficulty of labeling an instance for a human. Thus, the approach considers two things when selecting the next sample to clean: (1) how much the model disagrees with the label and (2) how difficult it is to clean for a human. To measure the latter, the authors considered the entropy of a model's predictions (the higher the entropy, the more difficult labeling). Other uncertainty-based approaches \citep{ponzio2021w2wnet, https://doi.org/10.48550/arxiv.1911.00068, liu2020data, ke2019multi} do not assume the presence of a human expert for cleaning instances; hence, they must automatically handle the cleaned instances. The straightforward strategy is to drop all the flagged instances \citep{ponzio2021w2wnet, liu2020data}. As a result, the dataset is purged from every potentially corrupted instance. Instead of removing all potentially corrupted instances, \citet{ke2019multi} iteratively removes the instances with the lowest confidence and retrains the model on the cleaned dataset until the model's confidence for all instances is above a threshold. \citet{https://doi.org/10.48550/arxiv.1911.00068} builds a class-conditional noise matrix and suggests different methods for record pruning using that matrix. This approach uses the confidence of the model to estimate class-conditional noise. An instance might belong to another class if the model confidence for that other class is above the per-class threshold. The per-class threshold of a class $j$ is the model's average confidence over the samples labeled $j$ for the same class $j$. Having a threshold defined for every class allows the approach to be more robust to heterogeneous class probability distributions and class imbalance. When an instance is above the threshold for more than one class, the approach selects the class the model is the most certain about. The authors propose five methods to prune out records while considering class-conditional noise. One of them prunes out the records that are predicted to belong to another class (when using the per-class thresholds).

For the uncertainty-based approaches to be successful, two things must be taken into consideration. First, the model should not overfit the dataset, or else the mislabeled instances will be correctly predicted (with respect to the labeled class, not the true class) and mislabeled instances will have a high confidence score. Second, an accurate measure of the model confidence must be used. Prior work \citep{https://doi.org/10.48550/arxiv.1506.02142, https://doi.org/10.48550/arxiv.1610.02136} has shown that the softmax score should not be used to measure the model's confidence, since it only acts as a normalization factor.

The first problem can be addressed using techniques that are robust to label noise, for example, \citet{bernhardt2022active} used Co-Teaching \citep{https://doi.org/10.48550/arxiv.1804.06872} and BYOL \citep{NEURIPS2020_f3ada80d}. Another solution to prevent the ML model from predicting the mislabeled class of an instance is to stop the training process before the model overfits the data. \citet{kohler2019uncertainty} made the observation that the mislabeled records are learned later in the model's learning process compared to the correctly labeled ones, and, as a result, stopping the training early would enable the model to have low confidence on mislabeled instances. Leveraging that observation, \citet{ponzio2021w2wnet} proposed an approach to automatically stop the training process. After each epoch, the model's confidence for each instance's labeled class is recorded and clustered using the k-means algorithm with a \textit{k} value of 2. Hence, after each epoch, a cluster with the most certain records and another with the least certain records are generated. When less than 1\% of the records change the cluster from one epoch to another, the model stops training. 

The second problem can be addressed using modeling techniques able to provide uncertainty scores, such as Bayesian Neural Networks (BNN) \citep{ponzio2021w2wnet}. The output of these NNs is a posterior distribution over the classes. Therefore, BNNs give a good measure of confidence when adequately trained. They are, however, complicated to implement and hard to train due to the high number of hyperparameters and computational cost \citep{ponzio2021w2wnet}. Thus, uncertainty-based approaches use different approximations of BNNs when concerned with providing a good measure of confidence. Inspired by the work of \citep{kohler2019uncertainty}, which also leverages Deep Ensemble \citep{https://doi.org/10.48550/arxiv.1612.01474}, \citet{ponzio2021w2wnet} approximates BNNs with Monte Carlo Dropout (MCD) \citep{Gal2016UncertaintyID}. MCD provides confidence measures that are numerically tractable and easier to implement than BNNs. 

\subsubsection{Loss-Based Approaches}
Similar to uncertainty-based approaches, loss-based approaches posit that a model trained on a dataset will have learned the representative features of each class and, as a result, will generate predictions that are not aligned with incorrect labels. However, instead of relying on a model's confidence, loss-based approaches will detect mislabels using the loss of a model. Records for which the model has a high loss are considered more likely to be mislabeled than other records. 

The authors in \citet{huang2021mislabeled, li2021research} first train a model on a dataset, then flag any record with a high loss value as mislabeled. Flagged records are then relabeled with the model's predictions. A new model is then trained on the modified dataset, and the process is repeated using the new model and the modified dataset. The authors in \citet{huang2019o2u} made the observation that NNs tend to overfit on noisy examples later in the training process than clean ones. Thus, corrupted examples are identified by looking for examples with a higher loss than average during training. However, the loss values of noisy and clean samples might vary depending on how the NN is initialized, which makes the approach less reliable. To address that concern, model training is repeated several times and the loss values are averaged. The weights of the NN are periodically re-initialized by suddenly increasing the learning rate and taking large gradient steps, which makes the model unlearn. The authors in \cite{yu2020unknown} address the challenge of detecting records not belonging to any of the classes of a dataset (noisy class). The approach adds a new class to the dataset for these samples. Then, a probability of 0.5 is assigned to the labeled class and the noisy class for every sample of the dataset. The approach iteratively trains a model and modifies the labels in such a way that the model's loss is decreased. Similarly to \citet{huang2019o2u}, the authors use a high learning rate to avoid overfitting on noisy labels. Similarly to Picket \citep{liu2022picket} (see Section \ref{fc:autoencoder}), the authors in \citet{zhang2019combining} measure the reconstruction error of autoencoders to detect corrupted samples. Examples with a high reconstruction error are flagged as mislabeled instances. Contrary to Picket \citep{liu2022picket}, \citet{zhang2019combining} trains an autoencoder for each class. To repair a mislabeled instance, they measure the reconstruction error of all the autoencoders for that sample and assign the class associated with the autoencoder with the lowest error. Similarly, the authors in \citet{salekshahrezaee2021reconstruction} used and compared three ML techniques, namely principal component analysis (PCA), independent component analysis (ICA), and autoencoders for unsupervised label noise detection. With each technique, they constructed a binary classifier that identifies instances with a high reconstruction error as anomalies. Based on their results, the autoencoder-based classifier obtained the best label noise detection score.

\subsubsection{Counterfactual Approaches} \label{lc:counterfactual}
A reason to do data cleaning is to improve a model's performance \citep{hara2019data}. This implicitly assumes that data errors reduce a model's performance. Thus, by finding dataset repairs that improve a model's performance, it is possible to find corrupted instances. In this study, we observed two types of dataset repairs: instance removal (i.e., dropping a record from a dataset) and attribute modification (i.e., modifying a record's features values, or labels). Counterfactual approaches try to find or repair corrupted records by comparing the performance of a model trained on a dataset with the performance of a model trained on the counterfactual dataset (i.e., the same dataset with some features or labels modified). In the following, we describe how the counterfactual modifications (i.e., instance removal and attribute modification) are used to detect mislabels and how they are implemented. Then, we cover the metrics used to quantify the impact of a counterfactual modification on a model's performance.

The first category of counterfactual approaches detects mislabels by comparing the performance of a model when trained on a dataset including a sample with the performance of a model when trained on the same dataset without that sample. A naive approach would train the model twice, once with the sample and another time without it. However, this can quickly become intractable as the size of the dataset increases. To address that problem, the authors in \citet{dolatshah2018cleaning} used parallel computing. Two other techniques have been proposed to address this problem in other works. The first technique is to use ML algorithms to predict the impact of including a record in the training dataset of a model \citep{smyth2020training}. In other words, an ML model is in charge of predicting the change in validation loss when a record is added to the training dataset of another ML model. To achieve that, \citet{smyth2020training} records a model's validation loss at every step of the training process. Hence, each time a model learns from a sample, the model is evaluated on a validation dataset. The validation loss is averaged per sample and the average becomes each record's label. Using this new dataset, ML algorithms can be used to predict the change of validation loss for any sample added to a dataset. The number of samples used to train the model predicting deltas in validation loss is a subset of the whole dataset to clean. The second technique to approximate the impact of removing a sample from a model's training dataset is using influence functions \citep{suzuki2021data, wu2021chef, koh2017understanding}. Influence functions are a technique from robust statistics that can be used to estimate how much a model's parameters change in reaction to an infinitesimal change in a training point's weight in a dataset. The influence function implementation of \citet{koh2017understanding} can estimate the effect of removing an instance from a dataset needing only an oracle access to gradients and Hessian-vector products of the ML algorithms. However, it is theoretically sound only for convex-loss models. To address this limitation, \citet{hara2019data} designed a novel estimator that, similar to influence functions, can estimate the change in performance if an instance is removed from the training dataset for models trained with stochastic gradient descent. In an experiment where the instances with the worst influence on a model's performance are progressively dropped, \citet{hara2019data} have shown that the approach improves the model performance more effectively than \citet{koh2017understanding}. Similarly to \citet{hara2019data}, \citep{suzuki2021data} and \citep{wu2021chef} provide improvements in terms of efficiency to \citet{koh2017understanding}'s implementation of influence functions, so that it can be used more efficiently for label cleaning.

The second category of counterfactual approaches detects mislabels by comparing
the performance of a model when trained on a dataset including a sample with the performance of a model when trained on the same dataset with a modified version of that sample. In addition to detecting potentially corrupted records, this counterfactual approach also provides repair suggestions (the counterfactual sample). \citet{xiang2019interactive, zhang2018training} formulate the search for mislabeled instances as an optimization problem where the goal is to maximize the model accuracy on an error-free dataset, by correcting the training samples' labels. Both approaches can automatically correct mislabels given an error-free test dataset.

Except for one paper in our selected papers, the counterfactual approaches included in this review evaluated modifications to a dataset using a loss function. However, as pointed out by \citet{flokas2022complaint}, loss functions are not always a good objective function to optimize for, since influence functions can not find records corrupted by systematic noise. Indeed, removing a record corrupted with systematic noise will not significantly impact the model's performance, since other samples are affected by the same noise. Hence, records corrupted with systematic noise are not influential per se. To address this issue, \citet{flokas2022complaint} designed an approach that uses complaints (filed by a human) to find corrupted instances. Complaints can be filed regarding tuples (e.g., ``that record should not have this label'') or aggregate values (e.g., ``there should be a fewer number of records that have a label''). The counterfactual approach then addresses that complaint using influence functions to find instances that, once repaired, will address the complaint.

\subsubsection{Outlier-Based Approaches}\label{lc:outlier}
Outliers are records that are significantly dissimilar from other records, to a point that one can believe that they have not been generated by the same process (as the rest of the data) \citep{8786096}. Thus, to detect mislabeled instances, outlier-based approaches for label cleaning search for records that differ significantly from the other records from a class. In \citet{guo2018automated}, a record is compared against every other clean record from its class using cosine distance. If the average distance is too far, then the record is marked as dirty. Because this approach does not scale to classes with many records, \citet{lee2018cleannet} compares instances against class prototypes; a vector representing all the instances from a class. The class prototype is generated by encoding a subset of the most representative records from a class using self-attention. Self-attention allows the encoder to focus on the most important instances when generating the final embedding. Records whose cosine distance with the class prototype is above a selected threshold are flagged as dirty. The approach iteratively runs the cleaning technique and trains a model on the cleaned dataset. The first layers of the trained NN are used to generate embeddings for the data cleaning steps. Thus, every time the NN improves performance, the data cleaning technique uses improved embeddings. \citet{bagherzadeh2017label} proposes a method for label denoising based on Bayesian aggregation. This approach addresses the limitations of kNN-based approaches, which are not robust to high levels of label noise rate since they only have a local view of instances. A local view refers to considering the immediate neighbors or similar instances of a data point within a certain vicinity. In contrast, a global view refers to comparing a record to others that are not in its neighbor. The proposed approach combines both views to detect mislabels. The global view of data is obtained by measuring the distance of an instance to its class distribution or by evaluating the value of the probability density function of each class at an instance. The aggregation leads to a robust detection of instances with noisy labels even in the presence of high levels of label noise.

\subsubsection{Other Approaches}
In this section, we cover other label cleaning approaches that are not included in the previous categories.

\begin{itemize}
    \item \citet{ekambaram2016active}: The authors argue that records near the decision boundary are more likely to be mislabeled by a human because they are more difficult to label. Thus, the authors propose to review support vector examples of an SVM by a human. Because there might be a lot of samples to review, they prioritize the ones that are on the wrong side of the decision boundary of another model trained on the dataset devoid of the support vector records.
    \item \citet{su2021correcting}: Instead of detecting and repairing errors, they generate a synthetic dataset devoid of mislabels. To do so, the authors leverage the phenomenon of mode collapse \citep{google_mode_collapse} in Generative Adversarial Networks (GANs). They train a GAN to generate realistic images for a class. Because GANs are subject to mode collapse, they expect the synthetic dataset to be free of mislabeled instances, since the GAN will not have memorized the less common instances of a class (i.e., mislabels).
    \item \citet{veit2017learning}: Instead of detecting errors and repairing them in two different steps, the authors propose an approach that directly maps an example's label to the correct value. An NN is then trained to predict the correct label given an instance and its label. An identity skip-connection is between the input label and the output of the NN. Thus, its task is to map the input label to the correct value. Similar to \citet{https://doi.org/10.48550/arxiv.1911.00068}, the authors argue that label noise is often class-conditional. Thus, learning to map labels from one class to another is sensible.

    \item \citet{klie2022annotation}: This study evaluates existing methods for detecting annotation errors and inconsistencies in text. Inconsistencies in this context refer to instances that should be labeled in the same manner to indicate the same type but are assigned different labels instead. The authors re-implemented 18 such methods and assessed their performance on several datasets for text classification, token labeling, and span labeling. The experiments indicate that inconsistencies are more challenging to detect than annotation errors. 
\end{itemize}

\subsubsection{Evaluation metrics} \label{lc:metrics}
The metrics used to evaluate label cleaning methods are highly similar to the ones used in feature cleaning works. Since label cleaning can be evaluated like a classification problem, metrics such as accuracy, F1 score, precision, and recall are commonly used \cite{rottmann2023automated, flokas2022complaint, smyth2020training, zhang2019combining, rehbein2017detecting}. A large number of works measured the improvement of a downstream model to evaluate their approach, since improving the performance of a model is often the end goal of correcting label errors \citep{veit2017learning, su2021correcting, suzuki2021data, wu2021chef, dolatshah2018cleaning}. In addition to measuring the correctness of the predictions of their data cleaning approaches, label cleaning works also considered other aspects, such as the execution time \citep{flokas2022complaint, zhang2019combining} and memory consumption \citep{gemp2017automated} of their approach.

\subsubsection{Comparison of the Approaches} \label{lc:comparison}
All types of label cleaning approaches propose viable strategies to detect mislabels. However, if one’s goal is to improve ML performance, counterfactual approaches are an interesting option. Indeed, their use of influence functions enables them to directly identify the samples that are the most likely to improve the performance of an ML model. However, counterfactual approaches suffer from a few limitations. Like many other label cleaning approaches, they are not robust to systematic noise when the goal is to improve ML performance \citep{flokas2022complaint}. Indeed, if the goal is to detect the samples that are the most likely to improve ML performance, then the samples tainted by systematic noise will not be identified, since the systematic noise will not be removed from the dataset by cleaning only one sample. Hence, other works propose alternative ways to use influence functions. As discussed in Section \ref{lc:counterfactual}, the authors in \citet{flokas2022complaint} let the user provide custom objectives that are less likely to be affected by systematic noise than the loss of ML models. Instead of using influence functions to detect incorrect labels directly, the authors in \citet{teso2021interactive} use them to find counterexamples to cleaned samples (i.e., samples that must have an incorrect label given a cleaned sample). Similar work is done in \cite{zhang2018training}. It is important to note that other label cleaning approaches are not necessarily impermeable to systematic noise. As shown in \cite{ekambaram2016active}, uncertainty-based approaches (and loss-based approaches) are similarly susceptible to systematic noise in a dataset. Indeed, an ML model can learn systematic noise and have high confidence on noisy samples. In addition to having this limitation, a model may also learn any sort of noise by overfitting. We discussed mitigation techniques for this problem in Section \ref{lc:uncertainty}. Hence, while all types of approaches can be susceptible to systematic noise, the flexibility of counterfactual approaches allows them to define objectives less influenced by such noise.  
\\
Other limitations of counterfactual approaches stem from the selected implementation for influence functions. Previous works have highlighted the efficiency issues counterfactual approaches faced when using influence functions \citep{zhang2018training, suzuki2021data}. To make them computationally tractable, other works have made assumptions that impose limitations on the generalizability of their approach. For example, \citet{hara2019data} proposed an approach to detect mislabels that only works for models trained with stochastic gradient descent. Similarly, the authors in \citet{wu2021chef} assumed strong convexity on the models (e.g., logistic regression). Thus, because of the constraints of many counterfactual approaches, practitioners might be deterred from using them to clean the labels of datasets. On the other hand, uncertainty-based and loss-based approaches tend to be easier to implement and present fewer constraints. For example, the approach described in \citet{https://doi.org/10.48550/arxiv.1911.00068} detects mislabel solely by inspecting a model’s confidence distribution once trained on a dataset. Similarly, the authors \citet{huang2019o2u} propose an approach that identifies incorrectly labeled data based on the loss of the model during training. For both approaches, little to no effort is required to configure the cleaning approaches once the training procedure of an ML model is configured.

\vskip2em
\begin{small}  
\begin{tcolorbox}[enhanced, breakable, title=Label Cleaning: Summary and Future Directions]
\noindent\textbf{General existing approaches}: 
\begin{itemize}
    \item \textit{Uncertainty-based approaches}: Approaches relying on a model's confidence to detect mislabels. Records for which a trained model predicts a different class than the one that is labeled have a higher chance of being mislabeled than other records. 
    \item \textit{Loss-based approaches}: Approaches relying on a model's loss to detect mislabels. Similar to uncertainty-based approaches, records that are mislabeled are more likely to be incorrectly predicted by the model than other records (and have a high loss). 
    \item \textit{Counterfactual approaches}: Mislabels may hinder a model's performance. Thus, if the performance of a model improves after a record has been removed or relabeled, then the record is most likely mislabeled. 
    \item \textit{Outlier-based approaches}: Records that are incorrectly labeled might differ significantly from the other records in the labeled class. These approaches rely on that observation to detect mislabeled.
\end{itemize}
\vskip1mm
\noindent\textbf{Future Directions}: 
\begin{itemize}
    \item Develop efficient and accurate methods to estimate the impact of removing a record from a dataset on ML performance. Current solutions may consume a lot of resources \citep{suzuki2021data, hara2019data} or be limited to a specific type of model \citep{koh2017understanding}. Improving these techniques will allow for more performant counterfactual approaches (covered in Section \ref{lc:counterfactual}) while enriching the knowledge of ML explainability.
    
    \item Combine label cleaning methods using ensembling techniques for improved performance. In Section \ref{fc:ensembling-approaches}, we described approaches that combine feature cleaning tools using ensembling techniques. Reusing the same ideas but for label cleaning could be an interesting research direction.
    
    \item Develop accurate model confidence measure. As we covered in Section \ref{lc:uncertainty}, uncertainty-based approaches rely on a model's confidence to detect mislabels. However, using a model's softmax score is not a good measure of model confidence as previous research pointed out \citep{https://doi.org/10.48550/arxiv.1506.02142, https://doi.org/10.48550/arxiv.1610.02136}. Future works could develop techniques to better measure a model's confidence. 
\end{itemize}
\end{tcolorbox}
\end{small}

\subsection{Entity Matching}\label{em}
\begin{figure}[t]
 \centering
\includegraphics[width=0.85\textwidth]{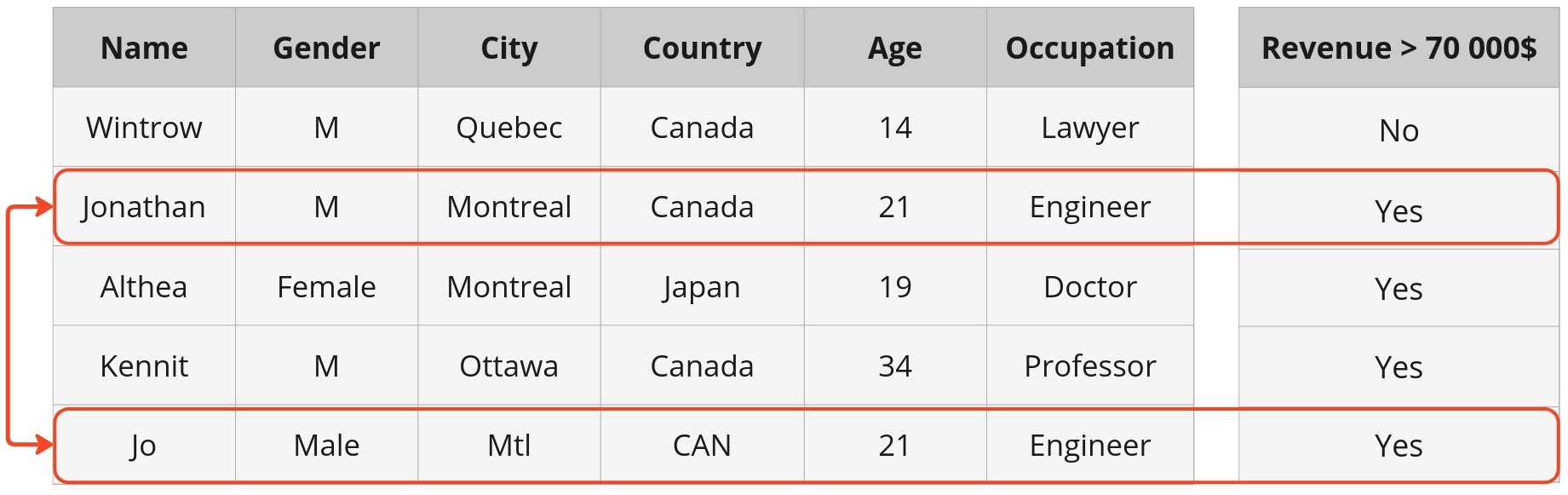}
 \caption{A table with duplicates (highlighted in red).}
 \label{fig:em_overview}
\end{figure}

In this section, we cover approaches to detect records referring to the same real-world entity. To better illustrate the type of error cleaned by entity matching approaches, we provide a table with duplicates in red in Figure \ref{fig:em_overview}. For ML, entity matching can be framed as a binary problem, where the model receives two records and must predict if they are the same entity or not. Formally, given a pair of records $(t_1,t_2)$, a model $M$ predicts if the pair of records is a match or not ($y \in [0,1]$). For conciseness, we use the word ``pair'' to refer to a pair of records. Furthermore, for simplicity, we call pairs referring to the same real-world identity as positive pairs, and negative pairs if they do not.

Deciding whether records refer to the same real-world entity essentially comes down to comparing records and looking for similarities and differences. Hence, when designing entity-matching approaches, researchers may bias the model into comparing records, which we refer to as the comparison bias. For example, \citet{jin2021deep} compare records by looking for shared and unique words between two records. The sets of shared and unique words are used by the model to decide whether they refer to the same entity or not. Because the comparison bias plays a central role in the design of the approaches, we grouped the approaches based on the way this bias is injected. Before the record-matching step, pairs that are most certainly not a match are filtered out of the dataset for efficiency purposes. This step is referred to as blocking. 

In the following, we explain entity matching and present the approaches included in our study in four parts. We will first cover the blocking strategies. Then we present the different entity-matching approaches using ML. Afterward, we show common techniques used to improve a model's performance on entity-matching tasks (e.g., data augmentation).  Finally, we describe how entity-matching tasks are evaluated in Section \ref{em:metrics} and provide a comparison of the entity-matching approaches in Section \ref{em:comparison}. Note that all the entity-matching approaches in our review operate over tabular data.

\subsubsection{Blocking} \label{em:blocking}
As mentioned earlier, blocking is applied before entity matching. The goal is to make the whole process of finding matching records more efficient. Instead of comparing each and every instance using the matching model, blocking places records into blocks so that entity matching is only applied to the tuples in the same blocks. This process effectively filters out pairs of records across different blocks, which have a low probability of being a match, making the whole process more efficient.

In the simplest case, hand-crafted rules can be used to filter out pairs of records that most likely are not a match. For example, for an entity matching task of companies' names, \citet{li2020deep} exclude pairs of records that do not share the same zip code or were not one of the top 20 nearest neighbors based on the TF-IDF cosine similarity on the name and address attributes. A downside of that approach is that designing blocking rules requires prior domain knowledge and can be time-consuming. Hence, \citet{ebraheem2018distributed} proposed an approach that is able to filter out pairs of records with minimal human involvement. To do so, Local Sensitive Hashing (LSH) \citep{LSH} is applied to the distributed representations (i.e., embeddings) of the tuples. Initially designed to efficiently approximate the k-nearest neighbors problem in sublinear time, LSH is used to filter out improbable matches. It hashes the distributed representations and uses the hash function to place each record into blocks, where each block holds records that are spatially close. Only pairs inside the same block can be considered for entity matching. \citep{ebraheem2018distributed} generate the records' distributed representations by transforming the record's tokens into embeddings using the Glove embedding model \citep{pennington2014glove} and merging them together with a bi-LSTM \citep{hochreiter1997long}. The pre-trained token embeddings are fine-tuned on the domain of the task using vocabulary retrofitting. Similarly, \citet{zhang2020autoblock} applies LSH on records' embeddings. To the difference of \citet{ebraheem2018distributed}, they generate more than one embedding per record. The records having at least one embedding in the same block as another record will be forwarded to the entity-matching model. The other pairs will be filtered out. The embeddings are generated by fusing fastText \citep{bojanowski2017enriching} embeddings using attention mechanisms. Similarly to this work, \citet{huang2023deep} uses the FAISS library \citep{johnson2019billion} to efficiently compute the k-nearest neighbors and then applies an algorithm to improve recall so that pairs of tuples that are a match are not filtered out. Essentially, the algorithm resizes the k-nearest neighbors set for each record (i.e., adjust ``k'') so that the likeliness of the k neighbor being a match is significantly more probable than the k+1 neighbor.

\subsubsection{Token Comparison}\label{em:token_comp}
The first category of entity-matching approaches compares records at the feature level. These approaches first extract a set of shared and unique words from the compared records, transform the words into embedding, and feed them to the model. \citet{jin2021deep} creates, for each attribute, a set of shared and different words. For example, for the values "Montreal, Canada" and "Quebec, Canada", the set of shared words would be "\{Canada\}" and the set of different words would be "\{Quebec, Montreal\}". Similarly, \citet{wang2020cordel} creates a set of shared words and two other sets containing the words unique to each record. The words in each set are then transformed into embeddings using a pre-trained language model \citep{jin2021deep} or word embeddings \citep{wang2020cordel}. Each set of word embeddings is then condensed down to an embedding using summation \citep{wang2020cordel} or attention mechanisms \citep{jin2021deep, wang2020cordel}. As a result, an embedding is created per set of words. These embeddings are then concatenated together and fed to the last layers of the model for prediction. In addition to the aforementioned strategies to combine embeddings, \citet{wang2020cordel} proposes to use the embedding resulting from the set of shared words to serve as the "query" key in the attention mechanisms applied to the sets of unique words. Their intuition is that information from the shared words is useful to know the important words in the set of unique words. 

\subsubsection{Latent Space Comparison} \label{em:latent}
The second type of entity-matching approach compares records in a latent space using a distance measure such as cosine distance before predicting whether a pair of records is a match or not. The records can be compared at the token level, the attribute level, or at the record level. We will refer to the embeddings used for comparison at each of these levels as token embeddings, attribute embeddings, and record embeddings respectively. Effectively, these approaches use one of the aforementioned embeddings for each record, compare the embeddings, aggregate the results of the comparison, and feed it to the entity-matching model.

\paragraph{Token-Level Comparison}
First, embeddings must be generated for each token. They can be generated from pre-trained embedding models \citep{fu2021hierarchical, mudgal2018deep}, transformers' encoders \citep{huang2023deep}, or any other suitable technique. To generate token embedding, \citet{li2020grapher} uses graph convolutional neural networks. The structured dataset is transformed into a graph by taking into consideration how frequent a word is for an attribute and the co-occurrence of words inside the same attribute. Once token embeddings are generated, they can be compared against the embeddings of the paired record. \citet{huang2023deep} compares each token embedding against the most similar token embedding from the other record and generates similarity (with scalar product) and difference vectors (with element-wise difference). Similarly, \citet{fu2021hierarchical} uses attention mechanisms to find the most relevant token to compare against and performs element-wise difference for comparing.  Instead of comparing word embeddings against each other, \citet{li2020grapher} compares the word embeddings of an instance against a representation of the other record's attribute. Attention mechanisms are used to generate the attribute representations and comparison is done with an element-wise difference. For every approach, after the token embeddings have been compared, they are condensed together in a unique representation for the record using an NN. The representation of each record is concatenated together and fed to the entity-matching model.

\paragraph{Attribute-Level Comparison}
When records are compared at the attribute level, embeddings must be generated for each attribute. \citet{mudgal2018deep, ebraheem2018distributed, kasai2019low} generate word embeddings for each word inside an attribute, then create an attribute embedding by merging the word embeddings together using variants of recurrent neural networks (e.g., LSTM \citep{hochreiter1997long}) or any other method (e.g., summation). \citet{nie2019deep} use attention mechanisms to strategically combine the records' token embeddings. \citet{bogatu2021cost, lattar2020does} use sentence encoders such as the universal sentence encoder \citep{cer2018universal} to directly output embeddings for attributes. Once the attribute embeddings are generated for both records of a pair, they can be compared using cosine distance \citep{lattar2020does}, or element-wise difference \citep{mudgal2018deep, ebraheem2018distributed, kasai2019low}. For every approach, the result of the comparison for each attribute is concatenated into a vector and fed to the last layers of the model for conducting the prediction task.

\paragraph{Record-Level Comparison} \citet{wang2022sudowoodo} generates record embeddings using a transformers' encoder. Tabular data is transformed into text using the serialization method of Ditto \citep{li2020deep}. The embeddings are compared using element-wise differences and fed to the model for the final prediction task. Because important information might be lost in the comparison vector (e.g., what the records have in common), they also give to the model the embedding of the concatenated records. 

\subsubsection{Learned Comparison} \label{em:learned}
The third category groups all entity-matching approaches that do not have a comparison bias and lets the model learn how to compare instances by itself. These approaches generally adopt superior modeling techniques, such as attention. \citet{mudgal2018deep} designed a NN with attention mechanisms and trained the model for entity-matching tasks. \citet{zhao2019auto} follows a three-step process to detect records referring to the same real-world entity. First, a model predicts the semantic type (e.g., company name, person name) for each attribute of a record. Second, an entity-matching model trained specifically for the semantic type predicted in the previous step is selected and predicts whether the attribute's value of both records refers to the same real-world entity (e.g., "Bill Gates" versus "William Gates"). The binary predictions (which are generated for each attribute) are fed to a one-layer NN to predict if the records are a match. As discussed in Section \ref{fc:transformer}, the authors in \citet{narayan2022can} evaluate LLMs on data cleaning tasks, notably entity-matching. To perform entity-matching, they first transform the tabular records into text format (see Section \ref{fc:transformer} for more details on the serialization procedure), show the serialized record to the model, then prompt it to predict whether the records refer to the same real-world entity or not. \citet{li2020deep, brunner2020entity} added a classification layer on top of a pre-trained transformer model and fine-tuned it for the entity matching problem. Similarly to the feature cleaning approaches with transformers (see Section \ref{fc:transformer}), both approaches transform tabular data into text format by concatenating the values of a record. Additionally, \citet{li2020deep} adds special tokens to differentiate the columns' names from the values of a record. For example, for a record of a person living in Montreal, Canada, a serialized entry could look like the following: [ATTRIBUTE NAME] city [ATTRIBUTE VALUE] Montreal [ATTRIBUTE NAME] country [ATTRIBUTE VALUE] Canada. In addition, authors added special tokens in the string for special types (e.g., phone number, zip code). 

\subsubsection{Other Approaches}
In this section, we cover other entity-matching approaches that are not included in the previous categories.
\begin{itemize}
    \item \citet{wu2020zeroer}: The authors propose an unsupervised approach for entity matching (i.e., no training data is required). Essentially, two Gaussian distributions are fit to the data, one to the positive pairs, and the other, to the negative pairs. The Gaussian distribution can then be used to estimate the likelihood of a pair of a record being a match or not. They propose optimizations to (1) reduce the number of parameters of the Gaussian distributions, (2) improve the prediction accuracy of their approach, and (3) address the performance issues caused by features with zero variance (i.e., the singularity problem in pattern recognition \citep{bishop2006pattern}).
    \item \citet{gottapu2016entity}: Instead of predicting if two records refer to the same entity, they predict to which entity each record refers. In other words, each class in their approach corresponds to a real-world entity. To classify records, they transformed each record into a sequence of word embeddings and then fed it to a convolutional NN for classification.
\end{itemize}

\subsubsection{Common Techniques to Improve Performance} \label{EM:dataset-improvements}
In this section, we cover various techniques commonly used to improve the entity-matching approaches' performance. Namely, these are data augmentation, semi-supervised techniques, active learning, and transfer learning.

\paragraph{Data Augmentation} 
Because entity matching is an unbalanced problem (i.e., there are more records that do not refer to the same real-world entity than do), generating negative pairs is fairly easy. Two records randomly selected from a dataset will most likely not refer to the same real-world entity. \citet{wang2022sudowoodo} randomly select two records that are similar but are not a match and assign them the negative label. The authors argue that selecting similar instances is superior to random selection since it makes the prediction problem more difficult for the model. Generating positive pairs is, however, more difficult than generating negative pairs. \citet{wang2022sudowoodo} applies task-specific data transformations that preserve the label of a pair of records. \citet{wang2022sudowoodo} picks any two records whose embeddings have a cosine distance inferior to a fixed threshold. We describe in Section \ref{em:learned} how the embeddings are generated. Instead of selecting instances, \citet{li2020deep, huang2023deep} apply transformations to a record that should preserve the record's identity (e.g., deleting an attribute's value). The modified record can replace the initial record in an already existing pair of records \citep{li2020deep}, or the modified record and the initial record can become a new pair \citep{huang2023deep}. To minimize the risk that the pair is not truly a match, \citet{li2020deep} proposes MixDA, a technique inspired by MixUp \citep{zhang2017mixup}. The approach interpolates the initial record embedding with its transformed version and adds the interpolated version to the training dataset.

\paragraph{Semi-Supervised Techniques}
To create positive pairs, \citet{kasai2019low} assigns a positive label to any pair that is rated as very likely to be a positive pair according to the entity-matching model. The model is then re-trained on these newly labeled samples and the process can be repeated. \citet{huang2023deep} used adversarial learning to generate additional training samples. The goal of the generator model is to fool the discriminator model (i.e., the entity-matching model) into believing that a generated record refers to the same real-world entity as a real record.

\paragraph{Active Learning}
\citet{kasai2019low} observed that entity matching is an imbalanced problem (i.e., there exist more negative pairs than positive pairs), and if it is not considered in the active learning process, more negative pairs than positive pairs might be labeled. Hence, the approach tries to send to the oracle an equal amount of positive and negative pairs by referring to the prediction of the entity-matching model, so only uncertain pairs are sent (i.e., when the probability that a pair is a match is close to 0.5). The oracle receives an equal amount of pairs slightly under and over the threshold (of 0.5). \citet{li2020deep}'s approach prioritizes samples with the highest informativeness. They start by grouping unlabeled pairs into clusters based on a custom similarity measure. A pair's informativeness is based on the number of samples that will be labeled if the current pair is labeled positively or negatively. \citet{meduri2020comprehensive} benchmarked different active-learning-based approaches for entity matching. They tried different combinations of models (i.e., linear, non-convex, tree-based, and rule-learning) with different acquisition functions to select pairs (i.e., query-by-committee \citep{freund1997selective}, how close a sample is from the decision boundary, and likely false positives/negatives \citep{qian2017active}). They adapt blocking (see Section \ref{em:blocking}) to sample selection in active learning to fasten the process. 

\paragraph{Transfer Learning}
Transfer learning refers to reusing a model trained on a different but related task hoping some of the knowledge will be transferable for the task at hand \citep{wikipedia_2023_transfer_learning}. To have better performances, some entity-matching approaches reuse the models trained for another entity-matching problem for the task at hand. For example, a model used to match citation records between two databases could be used for two other databases with different schema \citep{kasai2019low}. Here, we discuss the techniques used to make the transfer of knowledge learned from one task to another more efficient. As discussed in Section \ref{em:latent}, \citet{kasai2019low} predicts if two records are a match by comparing the records' tuple embedding. In order to maximize the benefits of transfer learning, they trained the model generating tuple embeddings to be dataset-invariant. In other words, the goal is for the embeddings from the target domain to be similar to the ones of the source domain. The authors use adversarial learning to train the embedding model to generate dataset-invariant embeddings. The role of the discriminator model in the adversarial setting is to predict the provenance of a record (i.e., to predict whether a record comes from the source or target domain). Similarly, \citet{jin2021deep} tries to increase domain adaptation of their approach (i.e., transfer learning) by minimizing the KL divergence between embeddings from the source domain with the ones from the target domain.

\subsubsection{Evaluation metrics} \label{em:metrics}
Similar to feature cleaning and label cleaning, entity matching can be evaluated like a classification problem. Hence, metrics such as accuracy, F1 score, precision, and recall are commonly used \citep{huang2023deep, li2020grapher, gottapu2016entity, wu2020zeroer, fu2021hierarchical}. The execution time of proposed approaches is sometimes presented, e.g., \citep{wu2020zeroer, mudgal2018deep, li2020deep}.

\subsubsection{Comparison of the Approaches} \label{em:comparison}
As for the feature cleaning approaches, the most performant entity matching methods use pre-trained transformer models. Latent space comparison approaches use these models to generate meaningful embeddings \citep{wang2022sudowoodo, huang2023deep}, while learned comparison approaches use them to perform entity matching directly \citep{li2020deep, narayan2022can}. In the work of \citet{brunner2020entity}, the authors experimented with various transformer architectures and compared their performance to “classical” DL methods in entity matching. On the most challenging datasets, the transformer models outperformed the other approaches by an average of 27.5\% for F1 score. Later works leveraging transformer models achieved even better results on various datasets \citep{narayan2022can, li2020deep, huang2023deep}. In addition to being more performant, these approaches require fewer training samples. Previous DL methods for entity matching require large amounts of samples, which are often unavailable in realistic scenarios \citep{kasai2019low}. In comparison, the prompting approach described in \citet{narayan2022can} surpassed “classical” methods for entity matching with only 10 training samples. A similar work which focuses on maintaining good performances while consuming fewer training samples is presented in \citet{wang2022sudowoodo}. By requiring fewer training samples to learn to do a task, training is faster for such approaches, sometimes up to the order of a magnitude \citep{brunner2020entity}. 
\\
While approaches leveraging transformer models tend to surpass others in terms of performance, they can be outperformed by simpler approaches. The authors in \citet{wang2020cordel} presented a token-comparison approach that obtained better results on selected datasets than other approaches leveraging transformer models \citep{huang2023deep, li2020deep}. As discussed in Section \ref{em:token_comp}, token-comparison approaches compare records by extracting a set of unique and shared terms between the compared records before feeding them to an entity matching model. To motivate their choice of following this approach, the authors highlight that embeddings generated by neural architectures tend to smooth out the differences between records that are vital for entity matching. Hence, for example, the values “6-foot sandwich” and “12-foot sandwich” could be incorrectly detected as the same entity because they are semantically close and have similar embeddings. This strength of token-comparison approaches can also be a double-edged sword, as the authors in \citep{huang2023deep} highlighted. Real-world entity matching datasets are dirty: data can be mistyped or misplaced in an incorrect table column \citep{mudgal2018deep}. Thus, finding matching tuples by comparing whether the tokens between two records are exactly the same is error-prone.

\vskip2em
\begin{small}  
\begin{tcolorbox}[enhanced, breakable, title=Entity Matching: Summary and Future Directions]
\noindent\textbf{General existing approaches}: The approaches can be grouped based on how the records are compared (for the purpose of entity matching). We identified the following three categories.

\begin{itemize}
    \item \textit{Token comparison}: A set of common and different words between two records are generated and fed to an ML model that will predict whether the records match or not. 
    \item \textit{Latent space comparison}: The records are compared in a latent space using a distance measure such as cosine distance. The distance measure can then be used by an ML model to predict whether the records match or not. The embeddings used for comparison can be token, attribute, or record embeddings.
    \item \textit{Learned comparison}: A model ingests both records and predicts whether the records match or not. Contrary to other approaches, the model is entirely responsible for learning how to compare records.
\end{itemize}
\vskip1mm
\noindent\textbf{Future Directions}: 
\begin{itemize}
    \item Explore how comparing records using different latent representations at the same time can improve entity matching. In Section \ref{em:latent}, we described three types of latent-space representation approaches (token-level, attribute-level, and record-level). Combining all three representation approaches could enable the detection of low-level differences between records as well as high-level ones, which could improve performances.
\end{itemize}
\end{tcolorbox}
\end{small}

\subsection{Outlier Detection}\label{sec:od}
In Section \ref{lc:outlier}, we presented approaches to detect mislabels using ideas from outlier detection. Here, we cover approaches to detect outliers in a dataset (not for the purpose of detecting mislabels). Borrowed from \citet{ilyas2019data}, we identified three categories of outlier detection approaches: statistic-based, distance-based, and model-based approaches. Note that an approach may belong to more than one category. For example, the approach in \citet{terrades2022flexible} uses a model to detect outliers (model-based), but also preprocess data before feeding it to the model using distance-based approaches. We describe how outlier detection tasks are evaluated in Section \ref{out:comparison}.

\subsubsection{Statistic-Based Approaches}
Statistic-based approaches detect outliers by looking for records that are in low-probability regions according to a stochastic model \citep{ilyas2019data}. To detect outliers, the authors in \citet{pit2016outlier} model a dataset using generative models and use them to detect outliers. Any record with a low probability according to the generative models is marked as an outlier. To improve the performance of their approach, the records are enriched with metadata (e.g., the string length of an attribute).

\subsubsection{Distance-Based Approaches} \label{outliers:distance}
Distance-based approaches measure the distance between a record and other records to detect outliers \cite{ilyas2019data}. For example, a record that is far from any other data points could be declared an outlier. \citet{guan2016wenn} compares a record against its neighbors to determine if it is an outlier. If a significant portion of its neighbors come from another class, the record might be an outlier. The authors argue that previous works often incorrectly detect legitimate records from a class with a low number of records as outliers since the records are often in regions with a lot of records from other classes. To mitigate this problem, a weighted average of the labels in the neighborhood of a record is performed to detect outliers and give a larger weight to samples that are close and that are from the minority class. Similarly, \citet{terrades2022flexible} uses the neighbor of a sample to determine if it is an outlier. The authors make the observation that outliers often are in clusters of samples with high heterogeneity of classes, or in clusters where a large proportion of the samples have the same class. Thus, to detect outliers, both aspects are measured and fed to a model for prediction.

\subsubsection{Model-Based Approaches}
Model-based approaches primarily rely on a classifier to detect outliers \cite{ilyas2019data}. Detecting outliers is formulated as a binary prediction task. \citet{liu2019generative} proposes an approach to detect outliers that is inspired by the GAN architecture \citep{goodfellow2014generative}. Their approach consists of a discriminator model and \textit{k} generator models whose goal is to generate synthetic samples. Each generator is trained on a different subset of similar data points to increase the diversity of generated samples. Once trained, the discriminator model can be used to detect outliers. The authors in \citet{jiang2023read} use the reconstruction error of an autoencoder (similar to autoencoder-based feature cleaning, see Section \ref{fc:autoencoder}) and the distance between a record and the closest class distribution to detect outliers (similar to distance-based approaches, see Section \ref{outliers:distance}). They conducted experiments with two distance measures, namely Mahalanobis and Euclidean distance. The more a record is poorly reconstructed and far from any class distribution, the more likely it is an outlier.

\subsubsection{Other Approaches}
In this section, we cover outlier detection approaches that were not included in the previous categories.

\begin{itemize}
    \item \citet{papastefanopoulos2021unsupervised}: Several unsupervised detecting approaches are proposed to detect outliers; however, according to the authors they rely predominantly on a single metric (e.g., distance, density, etc.), and each has its shortcomings. In order to address this issue, a voting ensemble-based approach of multiple outlier detection methods relying on different metrics is introduced. To improve performances, non-informative methods are filtered from the ensemble using an unsupervised spectral feature selection algorithm \citep{zhao2007spectral}.
    \item \citet{domingues2018comparative}: The authors conduct a study in which they compare outlier detection methods. The compared methods are grouped into probabilistic-based (e.g., robust kernel density estimators \citep{kim2012robust}), neighbors-based (e.g., Subspace outlier detection \citep{kriegel2009outlier}), distance-based (e.g., Mahalanobis distance \citep{ben2005outlier}), information theory-based (e.g., Kullback-Leibler divergence \citep{filippone2010information}), NN-based (e.g., Grow When Required (GWR) network \citep{marsland2002self}), domain-based methods (e.g., One-class SVM \citep{scholkopf1999support}) and isolation methods (e.g., Isolation forest \citep{liu2008isolation}). The study shows that isolation forest \citep{liu2008isolation} is an efficient method for detecting outliers in terms of average precision, training and prediction time, and scalability. 
    \item \citet{alimohammadi2022performance}: Similar to \citet{domingues2018comparative}, the work focuses on evaluating techniques for detecting outliers in time-series data. The evaluated techniques include ML-based techniques (e.g., one-class SVM \citep{scholkopf1999support}, statistical-based techniques (e.g., z-score \citep{rosner1983percentage} and regression-based techniques (e.g., polynomial fit \citep{motulsky2006detecting}. The experiments show that KNN is the optimal choice in terms of ML performance, simplicity of configuration, and computation time.  
\end{itemize}

\subsubsection{Evaluation metrics} \label{out:comparison}
Similar to feature cleaning, label cleaning, and entity matching, outlier detection can be evaluated like a classification problem. However, different metrics tend to be used to measure performance. It is common for outlier detection approaches to detect outliers by applying a threshold over a numerical value. Thus, to factor out the threshold configuration, metrics such as “Area Under the Receiver Operating Characteristic” (AUROC) \citep{jiang2023read, papastefanopoulos2021unsupervised, terrades2022flexible} are commonly used. The execution time \citep{alimohammadi2022performance, domingues2018comparative} and memory consumption \citep{domingues2018comparative} of the proposed approaches are sometimes presented.

\vskip2em
\begin{small}  
\begin{tcolorbox}[enhanced, breakable, title=Outlier Detection: Summary]
\noindent\textbf{General existing approaches}: 
\begin{itemize}
    \item \textit{Statistic-based approaches}: Records are marked as outliers if they are in a low probability region according to a probability model.
    \item \textit{Distance-based approaches}: The distance between a record and others is measured to determine if the record is an outlier.
    \item \textit{Model-based approaches}: A model plays a central part in determining if a record is an outlier.
\end{itemize}
\end{tcolorbox}
\end{small}

\subsection{Imputation}\label{imputation}
In this section, we cover approaches to impute values in datasets with missing values. \citet{abidin2018performance} compares the performance of three ML algorithms (i.e., decision trees, k-nearest neighbor, and Bayesian networks) for imputation on 10 datasets. They found that Bayesian networks are the most accurate for most datasets, but they are expensive to train. \citep{razavi2020similarity} proposes two techniques, kEMI and kEMI+ to impute categorical and numerical missing data. Both techniques first utilize the K-Nearest Neighbors (KNN) algorithm as a local search algorithm to find the K-top similar records to a record with missing values. Then kEMI invokes the Expectation-Maximization Imputation (EMI) algorithm which uses the feature correlation among the K-top similar records to impute missing values. The idea is to use conditional probability distribution of a feature given the other features using only the K-top similar records. In contrast, kEMI+ repeatedly invokes EMI to gather a collection of estimates for the missing value and combines them using Dempster-Shafer fusion \citep{ap1967upper,dempster2008upper}. Although kEMI+ can be more accurate than kEMI, it is less scalable. A novel technique proposed in \citep{silva2021co} that combines Adaptive Resonance Theory (ART) algorithm \citep{carpenter1991fuzzy} with Co-active Neuro-Fuzzy Inference System (CANFIS) \citep{tfwala2013prediction} to impute missing data. CANFIS integrates the reasoning capabilities of fuzzy systems and the computational capabilities of NN such that the fuzzy logic provides a linguistic representation of variables and performs fuzzy inference, while the NN performs computations on the fuzzy rule outputs and maps them to crisp output values. ART complements CANFIS by providing unsupervised learning and clustering of the input data which helps the fuzzy rules construction and enhances the model representation capabilities. 

\vskip2em
\begin{small}  
\begin{tcolorbox}[enhanced, breakable, title=Imputation: Summary]
\noindent\textbf{General existing approaches}: The approaches covered in our study impute missing data by using feature correlation or by combining NNs with fuzzy systems.
\end{tcolorbox}
\end{small}

\subsection{Holistic Data Cleaning} \label{holisitic-data cleaning}
In this section, we present data cleaning approaches that try to clean more than one type of error at a time. 

\subsubsection{Data Cleaning Pipeline Generation} \label{data cleaning-pipeline-generation}
Instead of directly cleaning data, the approaches in this category try to find the optimal sequence and configuration of data cleaning tools. The process of building the optimal cleaning pipeline can be formulated as a search problem. Given a metric to optimize, the goal is to find the optimal ordering and configuration of cleaning tools. These kinds of approaches are sometimes referred to as a cleaning optimizer \citep{neutatz2021cleaning}. The main factor that differentiates approaches from one another is the objective function and the optimization algorithm used, which we describe below.

Optimally, one would want to decrease the total number of errors in the dataset. However, that information is not always known a priori. Thus, a common approach when cleaning for ML is to use the performance of the model trained on the cleaned dataset as a proxy of data quality \citep{berti2019learn2clean, krishnan2019alphaclean}. If the ultimate goal of cleaning the dataset is to improve the model's performance, this objective can be preferred to any other proxy of dataset quality. \citet{li2019cleanml} showed that not all data cleaning tools improve a model's performance. Thus, cleaning a dataset using ML performance as a metric to maximize can avoid cleaning operations that decrease model performance. Other metrics can be used as well. For example, \citet{krishnan2019alphaclean}'s approach can minimize the number of outliers in a dataset or the number of integrity constraints violated.

Technically, ML hyperparameter optimizers, such as Python Hyperopt \citep{bergstra2013hyperopt}, could be used to configure the data pipelines. However, as \citet{krishnan2019alphaclean} pointed out, these optimizers do not leverage the incremental nature of data cleaning. Effectively, instead of evaluating the impact of a data cleaning tool on a model's ML performance right after it has been applied to a dataset, ML optimizers evaluate complete solutions (i.e., a fully configured data cleaning pipeline). To address that problem, \citet{krishnan2019alphaclean} uses beam search to find optimal pipeline configurations. The least promising pipelines are periodically pruned out based on the performance of the model trained on the current dataset. Another strategy proposed by \citet{berti2019learn2clean} is to use Reinforcement Learning (RL) to find the optimal configuration for data-cleaning pipelines. In their work, an agent iteratively selects a data cleaning tool that will be applied to a dataset. The change in a model's performance after applying the data cleaning tool serves as the reward function of the agent, and the state is the last cleaning tool applied to the dataset. In addition to the data cleaning tools, pre-processing operations, such as normalization, can also be selected by the agent.

Instead of searching for new pipeline configurations tailored to a dataset, \citet{gemp2017automated} reuses data cleaning pipelines that were shown to be effective on similar datasets. For this to be effective, a good measure of similarity between datasets must be used. The authors chose to represent datasets by vectors of 22 dimensions composed of common meta-features from the literature, such as mean feature skew. The datasets' vectors are compared against one another using L1 distance. While the results were inconclusive, they argue that their approach could open the path to new ways of generating data-cleaning pipelines.

\subsubsection{Other Approach}
\citet{li2019cleanml} benchmark different data cleaning techniques on real-world datasets and observe their impact on the performance of ML classification tasks. To perform their experiments, they used data cleaning tools from all the data cleaning categories discussed in this review (i.e., Section \ref{subsection:fc} to Section \ref{imputation}). Using different datasets, the authors measured the impact of using each tool individually on ML performance, then they repeated the experiments with different combinations of tools. Some of the findings of the study are that an improvement to a model's ML performance because of data cleaning will generalize to other models and that no single data cleaning tool is the best for all datasets.

\vskip2em
\begin{small}  
\begin{tcolorbox}[enhanced, breakable, title=Holistic Data Cleaning: Summary and Research Opportunities]
\noindent\textbf{General existing approaches}: 
\begin{itemize}
    \item \textit{Data cleaning pipeline generation}: Instead of directly cleaning data, these approaches try to find the best sequence and configuration of data cleaning tools to optimize a metric (e.g. ML accuracy of a model trained on the cleaned dataset). 
\end{itemize}
\vskip1mm
\noindent\textbf{Future directions}: We elaborate on future research directions for holistic data cleaning in Section \ref{fwr:holistic}.
\end{tcolorbox}
\end{small}

\section{Challenges in DC\&ML}\label{cha}
By conducting this review, we observed challenges spanning multiple data cleaning activities. In this section, we have a discussion about them hoping to help the research community address some of the most important ones. We provide in Table \ref{table:challenges_summary} a simplified description of the challenges listed here. As a result of this effort, we propose future work directions in Section \ref{sec:future-works}.

\renewcommand{\arraystretch}{1.5} 
 \begin{table}
 \caption{Summary of the challenges mentioned in Section \ref{cha}.}
 \centering
 \begin{tabular}{|>{\centering}m{3em}|m{30em}|}
 \hline
 \multicolumn{1}{|c|}{Section} & \multicolumn{1}{c|}{Challenges and problems}   \\
 \hline
  \multirow{1}{*}[-0.7em]{\ref{cha:datasets}}  
        & For many data cleaning activities, finding a labeled dataset is a challenging task, since they tend to be scarce and are rarely centralized in one place (e.g. with the use of a dataset repository). \\
    \hline    
  \multirow{1}{*}[-0.7em]{\ref{cha:imba}}  
        & Most data cleaning tasks are imbalanced problems: samples tend to be clean rather than dirty. As a result, labeled data cleaning datasets have only a few dirty (labeled) samples. \\
    \hline    
  \multirow{1}{*}[-0.7em]{\ref{cha:tooling}}  
        & We have not observed any actively maintained tool to allow practitioners to leverage the latest advances in data cleaning. \\
    \hline    
  \multirow{1}{*}[-0.7em]{\ref{cha:silo}}  
        & Real-world datasets may have more than one type of error. However, the vast majority of data cleaning approaches covered in this study  clean only one type of error.
        Additionally, there is little knowledge regarding how cleaning one type of error impacts how effectively other data errors can be cleaned. \\
    \hline   
    
\end{tabular}
\label{table:challenges_summary}
 \end{table}

\subsection{Scarcity of Data Cleaning Datasets} \label{cha:datasets}
To evaluate a data cleaning approach or train an ML model to be used in a data cleaning approach, one needs a labeled dataset. As building a labeled dataset is a laborious process \citep{gauen2017comparison}, researchers are interested in using the existing public datasets. Additionally, using public datasets provides a baseline for researchers to compare their approaches. The release of large public datasets has been known to propel research in ML \citep{gauen2017comparison}. For instance, the release of ImageNet datasets \citep{deng2009imagenet} led to the birth of new neural architectures such as AlexNet \citep{krizhevsky2017imagenet} and VGG \citep{simonyan2014very}. For data cleaning tasks, there exist popular datasets used across many papers such as the Hospital dataset, the Flights dataset, and the Beers dataset (all of which are described in \citet{mahdavi2019raha}). However, for researchers, finding and selecting a dataset to evaluate their data cleaning approaches can be a laborious process as it entails searching through many data cleaning papers for datasets. Furthermore, because data cleaning papers sometimes forget to provide the source of a dataset they used (and instead reference the paper they extracted the dataset from), obtaining the actual dataset can be arduous.

\subsection{Imbalanced datasets}\label{cha:imba}

In addition to the lack of public labeled datasets (Section \ref{cha:datasets}), the latter often only have a limited number of labeled dirty examples. As explained in Section \ref{fc:improving-dataset} and Section \ref{EM:dataset-improvements}, datasets of data cleaning activities are often imbalanced. The datasets used for feature cleaning generally only have a few dirty (labeled) records \citep{nashaat2021tabreformer, heidari2019holodetect}. Indeed, when cleaning the features of a dataset, one can expect to find a larger number of values that are correct rather than not. Hence, the cleaned versions of datasets have only a few labeled dirty samples. Similarly, for entity matching, most records in a dataset do not refer to the same real-world entity. Hence, generating "negative" data is easier than "positive" data since any pair of records sampled from a dataset is likely not to refer to the same real-world entity \citep{kasai2019low}.

\subsection{Lack of Proper Tooling}\label{cha:tooling}
Data cleaning is known to be tedious and time-consuming \citep{sambasivan2021everyone}. Researchers have reported that data cleaning can take up to 80\% of data scientists' time \citep{Press_2022}. However, as highlighted in \citet{sambasivan2021everyone}, data cleaning is one of AI's most under-valued and de-glamorized aspects. To clean data efficiently, practitioners might look for data-cleaning tools. If the practitioner wants the latest advances in data cleaning, he may look for the studies' replication packages. However, not all papers provide a replication package. Out of the 101 papers included in our study, only 20 provided one. Additionally, using replication packages is not convenient for practitioners. They must familiarize themselves with the application programming interfaces (API), which might not be user-friendly and differ from one replication package to another. Additionally, setting up the tool (e.g., installing dependencies) can be tedious and documentation may be lacking. Thus, to address this issue, \citet{muller2021papers} proposes OpenClean, an open-source library where all data-cleaning approaches can be centralized and made available via a unique API. Similar initiatives have gained a lot of traction in ML, such as Scikit-learn \citep{scikit-learn}. On the contrary, OpenClean has not received any support apart from its creators two years ago. Thus, a practitioner looking for data-cleaning tools might be deterred from using this open-source library.

\subsection{Siloed Data Cleaning Approaches}\label{cha:silo}
Except for the papers in Section \ref{holisitic-data cleaning}, the papers included in our study considered each data cleaning task individually. However, in the real world, practitioners find themselves cleaning more than one type of error in a dataset. For example, a practitioner could have to remove outliers, and then clean the features in a dataset. Cleaning each data error individually could lead to inefficiencies. For example, \citet{li2019cleanml} has shown that cleaning all data error types in a dataset does not necessarily improve a model's performance. Furthermore, cleaning all types of errors can lead to worse performance than cleaning only one type from a dataset. While understanding the interactions between the data cleaning approaches is important to effectively clean a dataset, they are not fully understood yet. The current approaches that clean more than one type of error in a dataset (Section \ref{holisitic-data cleaning}) do so by following the predictions of an ML model \citep{berti2019learn2clean} or by adopting a beam-search approach \cite{krishnan2019alphaclean}. They do not understand the influence the underlying data cleaning tools have on each other's performance and rather follow a search strategy guided by the impact the data cleaning tools have on a proxy of data quality.

\section{Future Directions} \label{sec:future-works}
In this section, we propose future work directions based on the challenges mentioned in Section \ref{cha} and the opportunities we have discovered in our study. Comparatively to the future directions proposed in Section \ref{sec:review} for each category in our taxonomy, the suggestions made in this section are applicable to all data cleaning tasks and do not stem from the idiosyncrasies of the respective problems. They are based on the challenges and opportunities shared between the tasks. We provide a summary of the research directions proposed in this section in Table \ref{table:fwr_summary}.

\renewcommand{\arraystretch}{1.5} 
 \begin{table}
 \caption{Summary of future work recommendations from Section \ref{sec:future-works}.}
 \centering
 \begin{tabular}{|>{\centering}m{3em}|m{30em}|}
 \hline
 \multicolumn{1}{|c|}{Section} & \multicolumn{1}{c|}{Research opportunities}   \\
 \hline
  \multirow{2}{*}[-0.7em]{\ref{fw:data-aug}}  
        & Develop approaches to find an optimal parametrization of data augmentation techniques.  \\
        & Develop approaches to automatically filter out invalid augmented samples.  \\
    \hline              
  \multirow{4}{*}[-0.7em]{\ref{fw:datasets}}  
        & Create datasets repositories for all data cleaning activities (e.g., feature cleaning).  \\
        & Build a taxonomy of data errors.  \\
        & Publish data cleaning datasets that are labeled.  \\
        & Facilitate the process of correcting errors in public ML datasets and keep track of the changes.  \\
    \hline              
  \multirow{1}{*}[-0.3em]{\ref{fw:tooling}}  
        & When proposing an approach, a researcher should implement its approach in a public data cleaning library. This should be enforced by publishers' reviewers. \\
    \hline              
  \multirow{3}{*}[-0.7em]{\ref{rec:LLMs}}  
        & Develop new data-cleaning pipeline generation approaches. \\
        & Develop approaches that can clean more than one type of error at a time. \\
        & Share LLMs pre-trained on data cleaning tasks. \\
    \hline              
  \multirow{4}{*}[-2em]{\ref{fwr:holistic}}  
        & Explore how LLMs can be leveraged for data cleaning tasks. \\
        & Explore how tabular data can be transformed into text for data cleaning with LLMs. \\
        & Develop tools combining data cleaning with other data pre-processing activities. \\
        & Analyze how different data cleaning activities affect one another when applied to the same dataset.
        \\
    \hline              
  \multirow{1}{*}[0em]{\ref{fw:idc}}  
        & Develop interactive data cleaning approaches. \\
    \hline              
\end{tabular}
\label{table:fwr_summary}
 \end{table}

\subsection{Data Cleaning Datasets}\label{fw:datasets}
To evaluate a data cleaning approach or train an ML model to be used in a data
cleaning approach, one needs a labeled dataset. However, as mentioned in Section  \ref{cha:datasets}, finding publicly available labeled data cleaning datasets can be a laborious task since there is not a central repository of such datasets for most data cleaning tasks. Previous work has led to the birth of a dataset repository for entity matching \citep{magellandata}. However, no studies included in our work mentioned such repositories. Labeled datasets are of paramount importance since they provide a benchmark to compare different approaches in addition to enabling the development of  ML approaches. Hence, a meaningful contribution to data cleaning would be to develop a dataset repository for the various data cleaning tasks covered in this study.

Providing a data repository would open up other new research directions. For feature cleaning, one could build a taxonomy of errors in ML datasets using the dataset repository. This taxonomy would help researchers evaluate their feature-cleaning approach in a systematic way. Instead of solely selecting datasets based on previous works, the researchers would be able to select datasets that have the error types their approach tries to address. 

Finally, in the same effort to provide data-cleaning datasets to researchers, future works could target publishing labeled data-cleaning datasets. To address the lack of labeled datasets, previous works \citep{flokas2022complaint, liu2022picket, neutatz2019ed2} have used datasets with artificial errors (i.e., errors injected into clean data). Thus, in an attempt to enable researchers to evaluate their approach against real-world errors, we encourage future works to publish labeled data-cleaning datasets. Additionally, maintainers of any public ML dataset could facilitate the process of identifying data errors and repairs, so as to organically create labeled data-cleaning datasets. 

\subsection{Data Augmentation}\label{fw:data-aug}
In Section \ref{cha:imba}, we described the issue plaguing many data cleaning problems: class imbalance. Indeed, labeled data cleaning datasets tend to have more clean (labeled) data than dirty (labeled) data. Hence, it is common for data cleaning techniques to leverage data augmentation to generate more training samples for the class that is underrepresented. While it is true that data augmentation can improve a model's performance, it can also reduce it, if not properly tuned \citep{miao2021rotom, wei2019eda}. For example, a data augmentation operator could transform a record in a way that its initial label does not apply anymore. Thus, when using data augmentation strategies, it is important to ensure they are correctly parameterized. Past studies in computer vision have tried to design search algorithms to automatically find the best parametrization of data augmentation techniques \citep{cubuk2019autoaugment, lim2019fast}. Future works could continue in this direction and evaluate their approach for data cleaning tasks. While a good parametrization increases the chances that good samples are generated from the data augmentation tool, all samples are not guaranteed to be valid. \citet{miao2021rotom} proposes an approach to automatically filter out augmented samples that are invalid. Additionally, the approach automatically weights examples based on their likeliness to improve a model's performance. Thus, the model learns more from samples that are weighted more. While not a complete replacement for a good parametrization of data augmentation tools, this strategy makes configuring data augmentation tools less critical. We encourage researchers to develop approaches to automatically filter out augmented records that hinder a model's performance.

\subsection{Tooling}\label{fw:tooling}
Data cleaning is known to be tedious and time-consuming \citep{sambasivan2021everyone}. Researchers have reported that data cleaning can take up to 80\% of data scientists' time \citep{Press_2022}. However, as highlighted in \citet{sambasivan2021everyone}, data cleaning is one of AI's most under-valued and de-glamorized aspects. To clean data efficiently, practitioners might look for data-cleaning tools. However, as illustrated in Section \ref{cha:tooling}, we only discovered one data cleaning tool openly accessible to practitioners. Only a few papers provided replication packages, which are not a convenient alternative to a public library. Despite being an efficient vehicle to make research developments available to practitioners faster, the open-source library we discovered has not received any contribution for the last 2 years. Hence, we encourage future works on data cleaning to provide a ready-to-use implementation of their approach on open-source libraries such as OpenClean \citep{muller2021papers}. Additionally, we suggest considering contributions to open-source libraries when evaluating data-cleaning approaches so as to incentivize researchers to contribute.

\subsection{Large Language Models} \label{rec:LLMs}
We presented in Section \ref{fc:transformer} and Section \ref{em:learned} data cleaning techniques that used LLMs to clean data. These approaches generally obtained state-of-the-art results on their respective tasks. Recently, OpenAI released ChatGPT \footnote{\url{https://openai.com/blog/chatgpt}}, an LLM which has shown impressive performances on diverse tasks, from solving programming bugs \citep{surameery2023use} to passing the Bar exam \citep{CNET}. We believe that future work should continue proposing data cleaning approaches that leverage LLMs, so as to benefit from the improvements with these models and achieve state-of-the-art performances \citep{narayan2022can}.

As shown in Section \ref{fc:transformer} and Section \ref{em:learned}, \citet{narayan2022can} ask questions in natural language to an LLM in order to clean data. For example, \citet{narayan2022can} uses the following sentence to clean data: "Is there an error in attribute X" (where X is the name of an attribute), to which the model answers "yes" or "no". Before asking the question, the tabular record is presented to the model in a special format that we describe in Section \ref{fc:transformer} and Section \ref{em:learned}. Their approach achieved state-of-the-art performances. We believe that an interesting future research direction would be to explore and compare different ways tabular data can be formatted for data cleaning with LLMs. Similarly, we encourage researchers to explore and compare different ways to prompt an LLM for data-cleaning tasks. This could build upon the current knowledge in prompt engineering \citep{white2023prompt}. 

Finally, we encourage future works to train and share LLMs trained on data cleaning tasks. Other researchers and practitioners could use these LLM models for their data-cleaning problems. A pre-trained model could leverage the knowledge it gained from previous data cleaning tasks for any other one. Similar contributions have been made in other fields \citep{beltagy2019scibert, araci2019finbert}. For example, in finance, \citet{araci2019finbert} developed FinBert, a language model fine-tuned on finance texts.

\subsection{Holistic Data Cleaning} 
\label{fwr:holistic}
In Section \ref{holisitic-data cleaning}, we presented two approaches to perform holistic data cleaning \citep{berti2019learn2clean, krishnan2019alphaclean}. These approaches search for the best order and configuration of data cleaning tools. They propose an attractive solution to practitioners who must configure various tools to clean diverse types of errors in datasets. However, only 2 papers included in this study proposed an approach to address that situation. Thus, we believe there is space in the literature to propose other data cleaning pipeline generation (DCPG) approaches. Future works could build upon the work done in AutoML \citep{he2021automl}, since the latter shares many similarities with DCPG. Indeed, DCPG could be seen as a subset of AutoML, where instead of considering the whole ML pipeline (as with AutoML), the search algorithm only focuses on a specific step in the pipeline (data cleaning).

More broadly, we encourage researchers to develop tools that try to clean more than one type of error in a dataset. A previous study \citep{wang2022sudowoodo} has demonstrated how two different types of errors (i.e., duplicates and feature errors) could be cleaned following a similar strategy. The authors \citep{wang2022sudowoodo} were able to adapt their entity-matching approach for feature cleaning with only minor modifications. Potentially, other data cleaning activities share similarities that are still uncovered. Thus, designing a learning approach that cleans more than one type of error at a time could clean a dataset more effectively than using each type of tool independently.

On an even broader scale, future work may concentrate on developing tools combining data cleaning with other data pre-processing activities. For example, in Section \ref{holisitic-data cleaning}, we presented the work of \citet{berti2019learn2clean}, who combined other data preparation tasks (e.g., feature normalization and selection) with data cleaning techniques (e.g., outlier detection). \citet{tae2019data} proposed a framework that combines data cleaning with unfairness mitigation and poisoned sample removal. 

To effectively design tools to holistically clean a dataset, one must understand how these tools interact with one another. However, we found little knowledge about that in our review of the literature, as mentioned in Section \ref{cha:silo}. As we covered in Section \ref{holisitic-data cleaning}, \citet{li2019cleanml} studied how different types and combinations of data cleaning tools impacted the performance of a model trained on the cleaned dataset. Instead of only considering an ML model's performance, researchers could study how applying one type of data-cleaning activity affects the performance of another one. For example, it could be interesting to study how feature cleaning influences an entity-matching model. Maybe cleaning data errors would make entity matching easier. Additionally, future works could also replicate \citet{li2019cleanml}'s study but with a larger set of tools, since \citet{li2019cleanml} used only one tool for some of the data cleaning activities that were considered (e.g., label cleaning). This work could use some of the tools reviewed in this paper.

\subsection{Interactive Data Cleaning}\label{fw:idc}
Most of the papers covered in this SLR (92/101 papers) provide standalone data-cleaning solutions that do not need someone's efforts to function (except for dataset labeling). In other words, the approach automatically cleans data once sufficient training data is provided. However, in practice, human experts are often involved in the cleaning process, since it is usually impossible for machines to clean datasets flawlessly \citep{ilyas2019data}. We refer to cleaning data using an expert's feedback as interactive data cleaning. These experts can provide valuable input to the data-cleaning process by solving various kinds of uncertainties. For example, for feature error repair (see Section \ref{subsection:fc}), experts can validate the repairs proposed by the approach to ensure that no errors are introduced in the dataset. Considering the low number of interactive data-cleaning approaches covered in this SLR and their relevance in practice, we encourage the community to develop such solutions.

\section{Related Works}
In this section, we present works that are related to our study. In Section \ref{rel:lit-rev}, we cover studies that, like ours, review the literature on DC\&ML.  While, as mentioned in Section \ref{Methodology:scope}, we limited ourselves to data cleaning works for image, text, and tabular data, data cleaning is applied to other data types as well. For instance, in SE, data cleaning is applied to code datasets. Hence, we present, in Section \ref{rel:dc-se}, data cleaning works in SE. In Section \ref{appendix:robust-learning}, we provide an overview of label-noise robust learning, an alternative method to label cleaning for improving the performance of ML models in the presence of label noise. We also highlight the synergies and similarities between the two approaches.

\subsection{Literature Reviews in DC\&ML}\label{rel:lit-rev}
Previous works partially cover DC\&ML. The authors in \citet{ilyas2019data} explain data cleaning and describe different data cleaning activities (e.g., entity matching, outlier detection). The authors reserved a full chapter on ML-based techniques for data cleaning. However, their work is not as exhaustive (in terms of approaches covered) as the work presented in this paper, and has a broader scope: all of data cleaning (not limited to ML). The authors in \citet{neutatz2021cleaning} present their vision of a holistic data-cleaning framework for ML applications. They also provide an overview of recent DC4ML or ML4DC approaches. Contrary to our work, they only provide an overview of the topic and do not examine the different techniques systematically and in detail as we do in this work. Similarly, \citet{10.1145/3506712} review the relationship between data cleaning and ML (i.e., DC4ML and ML4DC) and present current challenges in data cleaning for ML applications. In addition, they describe robust ML approaches. Their work is different from ours because their goal is not to systematically review the literature. The authors in \citet{thirumuruganathan2020data} explain how DL techniques can be leveraged for cleaning data and present other studies that leverage DL to clean data. As a vision paper, it proposes some research directions to the community. In \citet{roh2019survey}, authors conducted a survey of data collection for ML. They argue that data collection is composed of different processes, which feature cleaning, label cleaning, and entity matching are part of. Similar to ours, the data cleaning approaches presented are part of ML4DC or DC4ML. Contrary to our work, their scope is broader and they are not performing a systematic literature review. In the extended version of their study \citep{whang2023data}, in addition to covering data collection for ML, the authors cover robust model training and fair model training. Similar to the two previous works, a survey on data management for ML is reported in \citet{chai2022data} and includes a section to describe data cleaning for ML. \citet{dong2018data, laure2018machine, chu2016data} are tutorials that briefly discuss how ML can be used for data cleaning.

Other related works focus on one of the data-cleaning activities covered in our review (e.g., feature cleaning, label cleaning, etc.). For example, \citet{christophides2020overview} reviews the literature on entity matching including a section on entity matching using ML. \citet{barlaug2021neural} performs a survey on neural networks for entity matching. Authors in \citet{johnson2022survey, 10066216, karimi2020deep} review techniques to handle labeling errors in ML datasets including label cleaning. Authors in \citet{8786096, boukerche2020outlier} perform a survey of outlier detection techniques and cover ML-based techniques. The works of \citet{lin2020missing, adhikari2022comprehensive} survey techniques to impute missing data and discuss how ML can be used to impute missing data.

\subsection{Data Cleaning and Software Engineering}\label{rel:dc-se}
Datasets are used in various SE tasks, and, like any other dataset, they require quality assurance. Liebchen and Shepperd reviewed data quality within the context of empirical SE \citep{10.1145/2972958.2972967}. Tasks in empirical SE include vulnerability prediction or software effort estimation. The authors concluded that there is a growing acknowledgment within the community, compared to their prior study in 2008 \citep{liebchen2008data}, regarding the importance of data quality, with increased efforts directed towards exploring techniques for automatically detecting (and sometimes repairing) issues related to noise. They advised researchers to prioritize data quality, as even minor details can significantly impact results obtained from the data. They acknowledged the necessity of establishing mechanisms for assigning quality labels and discontinuing the use of datasets with serious concerns, since using non-qualified data is not productive for researchers or practitioners and wastes resources. Moreover, they highlighted that proposing protocols for describing data and addressing quality issues is gaining traction in the SE community. To ensure trustworthy data-driven research in SE, the authors clarified that having confidence in data quality and minimizing noise is crucial. Finally, they advised collaboration with practitioners in the collection and interpretation of data, as it will ultimately lead to more reliable and actionable research.
Bosu and MacDonell proposed a taxonomy of data quality challenges in empirical SE \citep{bosu2013taxonomy}. The authors reviewed the literature on data quality within empirical SE to pinpoint various quality challenges. They categorized these challenges into three primary groups. The first group encompasses data characteristics that render observations unsuitable for model learning (accuracy), like noise, outliers, and inconsistency. The second group includes challenges regarding the applicability of a model to different datasets (relevance), for example, heterogeneity and amount of the data. Lastly, the third group comprises factors restricting data accessibility and trustworthiness (provenance), like the commercial sensitivity of a dataset. Their goal was to utilize this taxonomy to highlight to the broader empirical SE community the challenges inherent in modeling with data and the proposed methodologies for identifying and resolving such challenges.

Bhandari et al. conducted a systematic literature review on data quality issues in software fault prediction \citep{bhandari2023data}. The authors reviewed 145 research papers published prior to November 2021 and investigated 5 aspects: data quality issue, pre-processing technique, modeling technique, dataset, and performance measures. Their results suggested that various data quality concerns, including data dimensionality and class imbalance, have received significant attention in existing SE literature for fault prediction. Nonetheless, issues such as class overlapping and missing data are particularly relevant to fault prediction datasets and warrant additional investigation. The comparative impact of addressing different data quality concerns remains largely unexplored. While C4.5, naive Bayes, multilayer perceptron, support vector machine, and random forest are commonly employed classifiers, their sensitivity to specific data quality issues should be considered when employing them. Accuracy, precision, recall, and area under the curve are frequently used performance metrics in fault prediction. The authors concluded that the presence of data quality issues in fault prediction datasets adversely affects classifier performance, highlighting the need for further research in this area.

In reaction to the previous studies describing the data quality issues of software engineering datasets, other works proposed data cleaning approaches to correct dirty data. A number of these works are dedicated to cleaning software vulnerability datasets. Croft et al. \citep{croft2023data} attempted to understand the characteristics of data quality in software vulnerability datasets. Authors measured the prevalence of each attribute in existing datasets and provided a discussion on the root causes of identified issues. They reported that, in real-world vulnerability datasets, 20-71\% of labels were not accurate, leading to a decrease (up to 65\%) in the performance of the trained models. Guo and Bettaieb \citep{guo2023investigation} define three critical issues (i.e., data imbalance, low vulnerability coverage, and biased vulnerability distribution) and three secondary issues (i.e., errors in source code, mislabeling, and noisy historical data) that can affect the model performance when trained on software vulnerability datasets. Croft et al. \citep{croft2022data} discussed the commonly used approaches for dataset cleaning for the software vulnerability prediction task. They include removing irrelevant code files, replacing the user-defined variables and function names with generic tokens, removing comments, blank lines, and non-ASCII characters, and removing duplicate code modules.

Instead of cleaning software vulnerability datasets, Liang et al. \citep{liang2023cupcleaner} introduced a data cleaning method tailored for the comment updating task, which is a burgeoning SE endeavor focused on automatically updating comments to reflect changes in source code. Their proposed solution, called CupCleaner (Comment UPdating’s CLEANER), is a semantic and overlapping-aware approach. CupCleaner calculates a score by considering both the semantics and overlapping information between the code and comments. By analyzing the score distribution, it identifies and filters out data with low scores located at the tail end of the distribution, effectively removing potentially unclean data. Similarly, the authors in \citet{shi2022we} developed a tool to automatically comment code after having conducted an analysis of noisy data in code comments. They evaluated their approach on 4 code summarization datasets.

While data cleaning can be applied on the datasets of various software engineering tasks, it is also interesting to note that a few of these tasks have similarities with data cleaning. For example, fault localization and fault repair seek to locate and fix errors in code \citep{10.1145/3631974}, which is similar to detecting and repairing errors in data (i.e., data cleaning). However, there are some differences between these tasks as well. First of all, artifacts are different: while the focus of data cleaning activities is on the data (in a dataset), fault localization/repair tasks are concentrated on the code. The symptoms of errors in data and code are different as well. Poor quality of the data leads to poor performance (in terms of prediction accuracy) of the ML model trained on it \citep{BRAIEK2020110542}, however, in software faults we observe compilation/runtime errors, memory issues or software failure \citep{7390282}. Moreover, evaluation metrics are sometimes different as we discussed in detail in Section \ref{sec:review}. Data cleaning activities are usually evaluated as a classification task, i.e., by measuring precision, recall, and F1 score. In some cases, the impact of data cleaning is measured on a downstream ML model. While evaluating fault localization techniques shows similarities with data cleaning (in the precision of identifying the location of errors in code and data), the effectiveness of a program repair is usually evaluated by passing unit tests.

\subsection{Label-Noise Robust Learning} \label{appendix:robust-learning}

Similar to label cleaning, label-noise robust learning approaches try to improve the performance of ML models in the presence of noise. However, in order to achieve that goal, label-noise robust learning approaches modify the learning algorithms instead of fixing data errors. Although data cleaning and robust learning are distinct approaches, they have similarities, as robust learning sometimes involves identifying and mitigating the effects of mislabeled samples in the training model. For example, DL techniques may incorporate a noise layer into the network, which aims to identify potentially incorrect labels based on discrepancies between the model's predictions (derived from prior layers) and the actual labels, thereby fine-tuning the model’s predictions and reducing the impact of mislabeled data \citep{thekumparampil2018robustness}. Additionally, some strategies minimize the influence of mislabels by down-weighting samples likely to be incorrect, by minimizing the loss on clean validation data \citep{ren2018learning}. Other effective strategies include employing a teacher-student model (knowledge distillation) \citep{shi2021distilling} or enabling models to abstain from making predictions on ambiguous samples, thereby allowing the model to concentrate on clearer, more reliable training data \cite{zhang2023learning}. Conversely, other robust learning approaches do not focus on the samples but instead adjust hyperparameters, such as the learning rate and regularization parameters \citep{zhou2021learning}. Moreover, some classification methods are designed to be used for both; robust learning and data cleaning like the approach proposed by Zhang et al. \citep{zhang2018data}. This approach employs a separate autoencoder to learn the features of each class and then utilize minimum reconstruction error for identifying mislabeled samples or classifying data points.

The relationship between robust learning and data cleaning is synergetic. As explained in Section \ref{lc:uncertainty}, robust learning approaches can be used to improve the performance of label cleaning approaches.
While data cleaning seeks to directly eliminate or rectify mislabels, robust learning develops models that are inherently more tolerant to such noise. This can be achieved by identifying and compensating for mislabels during the learning process or through adjustments in the training pipeline.

\section{Threats to Validity} \label{sec:threats}
We use the threats to validity described in \citet{feldt2010validity, zhou2016map} to analyze the limitations of our works. We detected three types of threats to validity in our work: construct validity, internal validity, and conclusion validity.

\textbf{Construct validity.} First, there is a risk that our search query does not allow us to exhaustively cover the field of interest of this study, preventing us from effectively answering our RQs. As mentioned in Section \ref{Methodology.paper_search}, we did preliminary searches to build a list of keywords that most likely capture the majority of the papers we are interested in. However, there is a risk that we missed keywords that would have led us to papers relevant to our study. To mitigate that threat, we performed snowballing, as described in Section \ref{met:snowballing}. The second threat relates to the existence of papers relevant to the study but published after our data collection date, and hence, not included in this paper. Future works should consider replicating our study. There is also a risk that the academic databases used in this study might be incomplete, i.e., missing some papers relevant to our study. To mitigate this threat, we searched on Google Scholar, which allowed us to find papers from other databases (which we had access to but were not on our list). Another threat to construct validity relates to the risk that our inclusion and exclusion criteria might have excluded works relevant to our study. However, similar criteria have been used successfully in previous studies \citep{tambon2022certify}. Also, papers relevant to our study might have been excluded because they processed a different data type than image, text, or tabular data (i.e., the data types considered in our study). However, as mentioned in Section \ref{Methodology:scope}, we selected three of the most common data types. Thus, we believe our scope allowed us to cover the most relevant papers for our study.

\textbf{Internal validity.} There is a risk that papers relevant to our study have been incorrectly excluded from our study while (1) applying inclusion and exclusion criteria (Section \ref{inclusion-exclusion-criteria}) or while (2) applying quality control questions (Section \ref{quality-control}). To avoid excluding relevant papers, in the first filtering step, we included papers in case of doubt. As mentioned in Section \ref{inclusion-exclusion-criteria}, a second researcher reviewed the list to exclude any paper that should not be included. To avoid excluding relevant papers in the second filtering step, a second researcher read every paper excluded because of the quality control questions and verified that the exclusion of a paper is justified, as mentioned in Section \ref{quality-control}. There is also a risk that one of the authors misunderstood the approach proposed in a paper leading to an incorrect explanation of the paper in our review. To mitigate this threat, each reviewer asked other reviewers to read the paper and share their understanding whenever they felt they did not fully understand the approach. Furthermore, a researcher verified that the papers' descriptions in our review were coherent with the papers' abstracts.

\textbf{External validity.} External limitations of our study are related to the generalization of the presented results. We attempted to cover different topics in DC4ML and ML4DC as much as possible. Nevertheless, given the scope and selection criteria used in our study, we believe that our findings can be generalized.

\textbf{Conclusion validity.} There is a risk that our study can not be reproduced. To mitigate it, we provide a replication package containing the list of papers retained after each step in the filtering process described in Section \ref{sec:paper_selection}, along with their quality ratings (see Section \ref{quality-control}) and extracted data (see Section \ref{sec:data_extraction}) when applicable. Any script used to help us conduct the study is also provided in the replication package.

\section{Conclusion} \label{sec:conclusion}
This study provides a comprehensive systematic review of the literature on DC4ML and ML4DC, i.e., data cleaning with and for ML. Our study includes 101 papers published between 2016 and 2022 inclusively, from various academic databases. We covered 6 different types of data cleaning activities: feature cleaning, label cleaning, entity matching, outlier detection, imputation, and holistic data cleaning. Based on our review, we provide 24 future research directions and we highlight some exciting data-cleaning approaches that can be further extended. Our future research directions revolve around 6 core ideas: (1) to improve data augmentation techniques for data cleaning, (2) to create more public data cleaning datasets, (3) to provide better tooling for cleaning datasets, (4) to study how LLMs can be used for data cleaning, (5) to explore holistic data cleaning approaches, and (6) to develop more interactive data cleaning approaches. We hope that this paper will serve as a solid foundation for future works on this important topic. 

\begin{acknowledgements}
This work is funded by the Fonds de Recherche du Quebec (FRQ), the Canadian Institute for Advanced Research (CIFAR), and the Natural Sciences and Engineering Research Council of Canada (NSERC). We would like to thank Dr. Hyacinth Ali for contributing to improving this SLR with his valuable comments.
\end{acknowledgements}

\section*{Declarations}
\subsection*{Data availability}
All data generated or analyzed during this study are available in the GitHub repository to help reproduce our results \citep{replicationpackage}.

%
\subsection*{Competing interests}
The authors declare that they have no conflict of interest. 



\bibliographystyle{spbasic}      
\bibliography{references}   

\clearpage



\end{document}